%% file: main.tex
\documentclass{cameravlm}
\input{math_commands.tex}

\usepackage{hyperref}
\usepackage{url}
\usepackage{booktabs}
\usepackage{graphicx}
\usepackage{amsmath}
\usepackage{xcolor}
\usepackage{colortbl}
\usepackage{rotating}
\usepackage{xspace}
\usepackage{amsthm}
\usepackage{capt-of}
\usepackage{placeins}

\newcommand{\method}{CameraVLM\xspace}

\title{When Left Is Right? Camera-Motion \\ Understanding beyond Screen}

\author[1,2]{Chong Cheng}
\author[2\ddagger]{Ran Zhang}
\author[2]{Xinyuan Li}
\author[2]{Alan Zhao}
\author[1\dagger]{Hao Wang}
\author[3]{Xi Wang}
\affiliation[1]{The Hong Kong University of Science and Technology (Guangzhou)}
\affiliation[2]{Tencent\\}
\affiliation[3]{LIX, Ecole Polytechnique, IP Paris}
\abstract{
Can a vision--language model tell left from right? Modern VLMs solve visual puzzles and follow hour-long videos, yet when asked whether the camera moved left or right, most do no better than a coin flip. Adding camera-motion data or geometric representations has been tried, separately, and neither closes the gap. We argue that camera-motion understanding needs two conditions at once: the representation must couple off-screen camera motion with on-screen flow dynamics, and it must be interpreted by the language model rather than trained as a standalone pose or depth estimator. We propose \method{}, which pairs a motion-state representation with state--language alignment. Camera tokens capture how the camera moves off screen, flow tokens capture how the subject and background move on screen, and a two-stage training strategy keeps language in the loop so that both are learned together with camera-motion semantics. With only 9B parameters, \method{} outperforms all open-source and camera-motion-specific models by a wide margin across three benchmarks and more than 20 baselines, and matches or surpasses GPT-6 Astra as well as human viewers in a 48-participant study. Project page: \url{https://3dagentworld.github.io/CameraVLM/}.

}
\begin{document}
\raggedbottom

\maketitle
\thispagestyle{firstheader}
\begingroup
\renewcommand{\thefootnote}{\fnsymbol{footnote}}
\begin{NoHyper}
\footnotetext[2]{Corresponding author.}
\footnotetext[3]{Project leader. This project was jointly supervised by Hao Wang and Xi Wang.}
\end{NoHyper}
\endgroup

\begin{center}
\includegraphics[width=\linewidth]{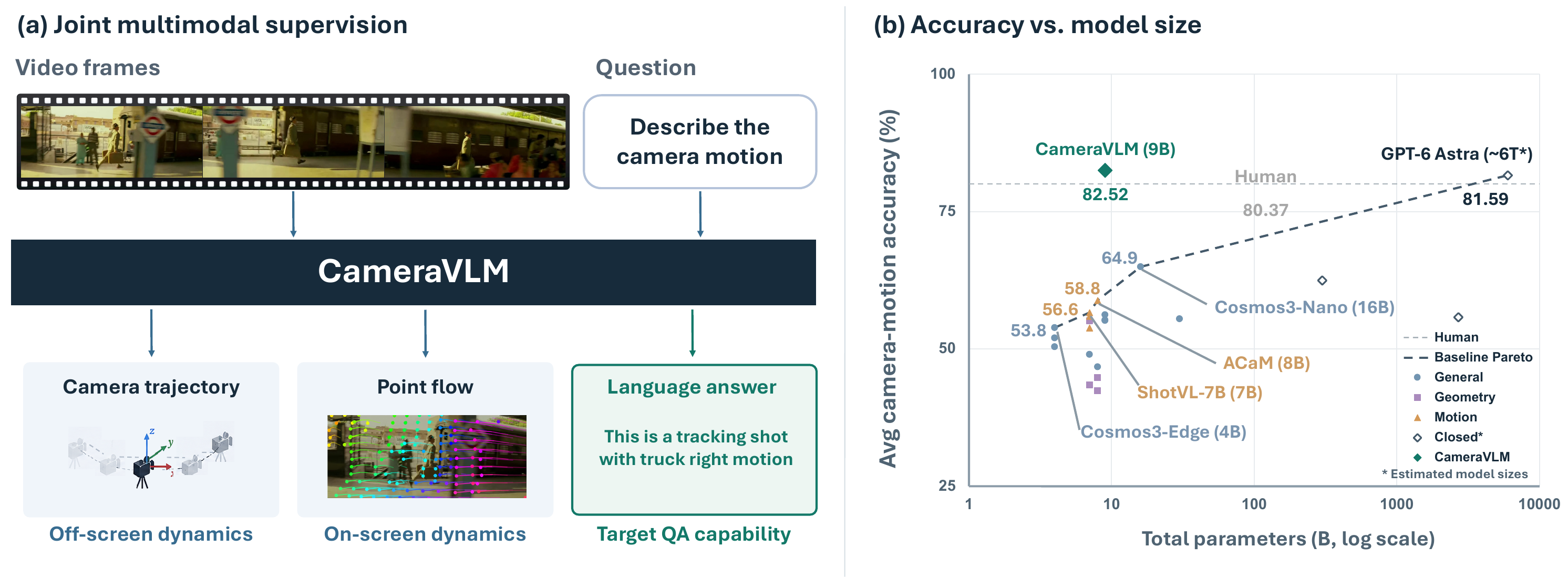}
\vspace{-20pt}
\captionsetup{hypcap=false}
\captionof{figure}{\textbf{\method{} learns camera motion from off-screen and
on-screen dynamics.}
(a) Given video frames and a question, \method{} is trained to recover the
camera trajectory (off-screen dynamics) and the point flow of the scene
(on-screen dynamics), and answers the question in language from these states.
(b) Average understanding accuracy on CameraBench, CineTechBench, and ShotBench
versus total parameters. With 9B parameters, \method{} lies above the Pareto
front of existing models and surpasses GPT-6 Astra and the human average; sizes
of closed-source models are estimated.}
\label{fig:overview}
\vspace{-6pt}
\end{center}

\section{Introduction}
\label{sec:intro}

\begin{figure}[t]
\centering
\includegraphics[width=\linewidth]{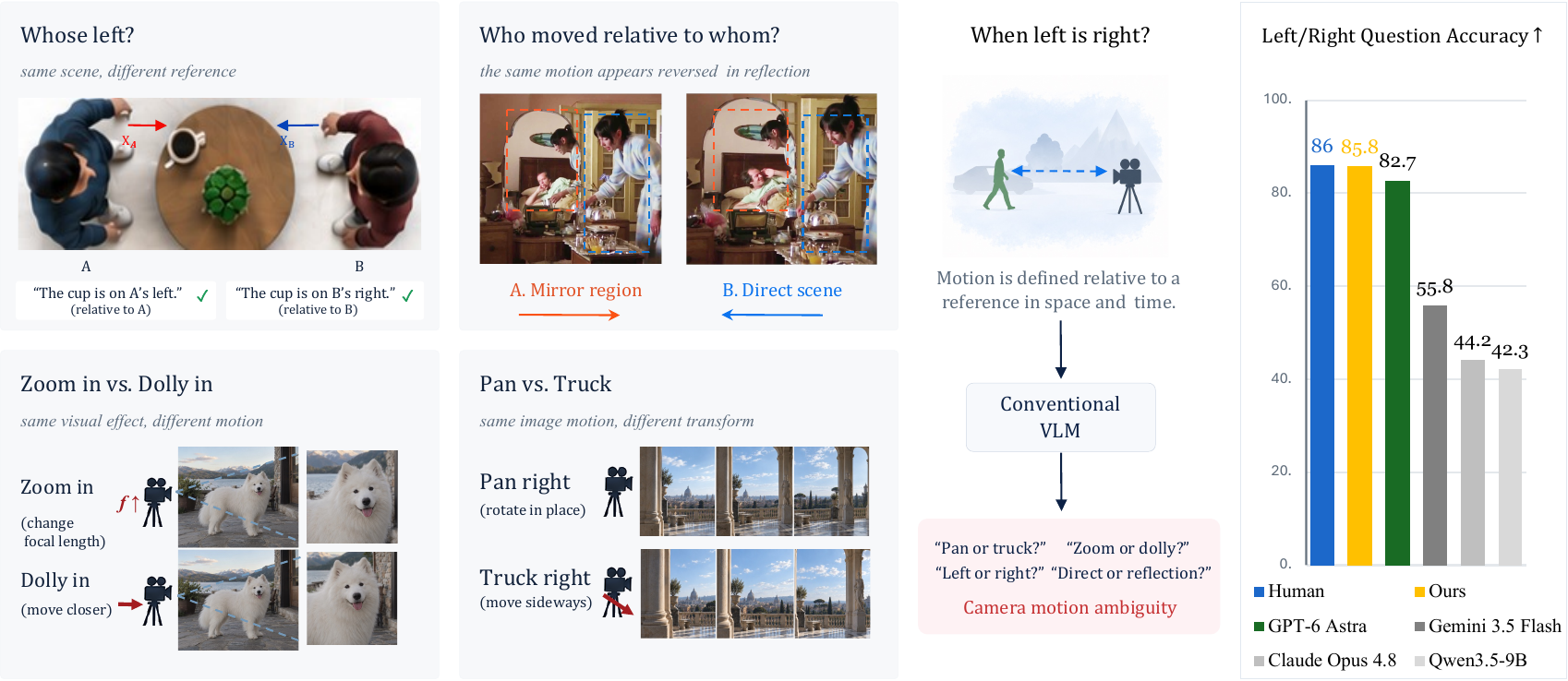}
\vspace{-22pt}
\caption{\textbf{Spatial motion semantics depend on reference and spatio-temporal state.}
The same scene can yield opposite directions under different references or similar appearances from different camera motions. \method{} models these relations through camera pose and flow. On left/right questions, most existing VLMs still perform poorly, while \method{} approaches human performance.
}
\vspace{-15pt}
\label{fig:teaser}
\end{figure}

Can Vision Language Models tell left from right? The answer seems obvious. Modern Vision Language Models (VLMs) such as Qwen3-VL~\citep{qwen3vl_tr} solve visual puzzles~\citep{visualpuzzles}, carry out multi-step visual reasoning~\citep{mathvista,mmmu}, and answer questions about hour-long videos~\citep{videomme,videochatflash,internvideo25}, and they increasingly serve as the perception and reasoning core of embodied agents~\citep{rynnbrain,hyembodiedvlm,cosmos3}. Yet ask whether the camera in a video moved left or right, and most open-source and commercial VLMs are correct only about half the time, no better than flipping a coin~\citep{camerabench,lrrbench}, while humans reach 86\% (Fig.~\ref{fig:teaser}). Is this failure caused by a lack of camera-motion data, a lack of geometric representations, or something else?

Both have been tried, separately. One line of work adds data: CameraBench~\citep{camerabench} and ACaM~\citep{acam} improve camera-motion recognition through motion-specific data and instruction tuning, mapping videos to fixed motion labels. The other focuses on representations: DepthLM~\citep{depthlm} and DepthVLM~\citep{depthvlm} model depth and scene structure, while Cambrian-P~\citep{cambrianp}, VLM-3R~\citep{vlm3r}, and CamInject~\citep{camdistill} incorporate camera pose or 3D geometric features. Neither is enough on its own (Table~\ref{tab:main}): camera-motion-specific models improve on the benchmark they are tuned for but often fail to transfer to others, and models with explicit geometry perform no better than general VLMs of the same size, some even worse. In this paper, we argue that this is because camera-motion understanding requires two conditions to hold together, and neither direction provides both: \emph{the representation must couple off-screen 3D geometry with on-screen image motion rather than encode geometry alone, and it must be reasoned over by language rather than merely trained as a pose or depth estimator}.

First, camera motion is observed through its interaction with scene content.
Camera translation and rotation can produce similar \textbf{on-screen} changes, while
subject motion may dominate the apparent motion within the frame. At the same
time, limited on-screen motion does not imply a stationary camera relative to
the surrounding or off-screen scene.
The same camera move looks different depending on what is in front of it: a dolly toward a distant scene barely changes the image, and a pan and a truck are hard to tell apart unless the scene has both near and far objects.\footnote{In cinematography, a \emph{pan} rotates the camera in place, like turning one's head; a \emph{truck} moves it sideways; a \emph{dolly} moves it toward or away from the scene.} The representation must therefore capture how the on-screen subject and background move in the image, together with the \textbf{off-screen} camera geometry. Second, spatial descriptions depend on a reference, and the question specifies it. When A and B face each other, a cup on A's left is on B's right; when the camera follows a walking person, the camera moves relative to the street but not to the person.
Camera-motion understanding therefore requires a representation that jointly
encodes off-screen camera motion and on-screen image motion, and that the
language model can interpret under the reference specified by the query.

We propose \method{} (Fig.~\ref{fig:overview}), which addresses these two conditions through motion-state representation and state--language alignment. On the representation side, we add two sets of tokens to the VLM's token sequence, processed by the backbone together with the video and question tokens. Pose-grounded camera tokens, whose hidden states are decoded into camera pose, encode the off-screen camera motion, namely its translation, rotation, and viewpoint change; flow tokens, pooled from the flow pathway, encode the on-screen motion, namely the cross-frame motion of points on the subject and background and whether they remain visible. Together, they let the model separate camera motion from subject motion and tell apart camera moves that look alike on screen.

On the alignment side, we construct joint training data with language, camera-pose, and flow supervision, and adopt a two-stage training strategy that keeps language in the loop, so the new states are learned together with their camera-motion semantics and remain interpretable by the language model. The first stage fits the new tokens to the language model: with it frozen, they are trained with pose, flow, and QA losses, and the QA gradient through the frozen model aligns them with its existing representation space. The second stage teaches the model to use them: the language model is unfrozen and trained jointly on language, pose, and flow objectives, with general video QA mixed in to preserve language abilities.

We evaluate on three camera-motion benchmarks (CameraBench, CineTechBench, and ShotBench) against more than 20 models, spanning open-source, geometry-aware, camera-motion-specific, and commercial VLMs, together with a 48-participant human study. With only 9B parameters, \method{} outperforms all open-source and camera-motion-specific models by a wide margin and, on average, matches or surpasses GPT-6 Astra and human viewers.

\begin{samepage}
Our contributions are as follows:
\begin{itemize}\itemsep2pt
\item We argue that camera-motion understanding in VLMs requires two conditions to hold together: a representation that couples off-screen camera motion with on-screen flow dynamics, and interpretation of that representation by the language model.
\item We propose \method{}, which pairs a motion-state representation of camera and flow tokens with state--language alignment through a two-stage training strategy that keeps language in the loop.
\item Across three camera-motion benchmarks and a 48-participant human study, \method{}, with only 9B parameters, outperforms all open-source and camera-motion-specific models by a wide margin and matches or surpasses GPT-6 Astra and human viewers on average.
\end{itemize}
\end{samepage}

\section{Related Work}

\paragraph{Camera-motion understanding.}
CameraBench~\citep{camerabench} evaluates video VLMs on camera-motion classification. CineTechBench~\citep{cinetechbench} and ShotBench~\citep{shotbench} broaden the scope to cinematographic attributes, while ACaM~\citep{acam} further decomposes camera motion into fine-grained atomic movements. Another line introduces explicit camera information through camera pose, camera tokens, or external motion estimates~\citep{cambrianp,camdistill,geoguided}. These methods improve camera-motion recognition but mainly operate on predefined motion semantics, without an explicit spatio-temporal state that can be reinterpreted as the target, reference, or temporal range changes.

\paragraph{Spatial representations for VLMs.}
DepthLM~\citep{depthlm} and DepthVLM~\citep{depthvlm} improve spatial representations through depth supervision. Cambrian-P~\citep{cambrianp} and VLM-3R~\citep{vlm3r} introduce camera pose, camera tokens, or richer spatial states, while SpaceMind~\citep{spacemind} uses camera representations to guide modality fusion and Q-GeoMem~\citep{qgeomem} further learns question-guided geometric memory. These methods mainly model depth, pose, or scene structure, with limited explicit modeling of cross-frame motion correspondence and continuous spatio-temporal states required for camera-motion understanding.

\paragraph{Embodied and world models.}
RynnBrain~\citep{rynnbrain_v1,rynnbrain} and Hy-Embodied-VLM~\citep{hyembodiedvlm} learn states for egocentric perception, action, and interaction~\citep{10972824}, while Orca~\citep{orca} and Cosmos~\citep{cosmos1,cosmos3} model broader physical states and state transitions. These models capture richer dynamics and physical information but do not explicitly resolve reference ambiguity in camera motion. The motion of the camera, subject, and background still changes with the reference entity and frame, so a richer dynamic state alone does not determine camera-motion semantics.

\paragraph{Camera and motion state estimation.}
Visual geometry methods predict camera and geometric states~\citep{dust3r,mast3r,vggt,cheng2026longstreamlongsequencestreamingautoregressive,cheng2026horizonstreamlonghorizonattentionstreaming}, while tracking methods estimate cross-frame trajectories~\citep{tapir,cotracker,cotracker3,alltracker,cowtracker}. VGGT4D~\citep{hu2025vggt4dminingmotioncues} separates moving objects from static backgrounds to improve pose and geometry estimation. Camera geometry also supports Gaussian scene reconstruction~\citep{cheng2025graphguidedscenereconstructionimages,cheng2025reggsunposedsparseviews,cheng2025unposed3dgsreconstructionprobabilistic}, outdoor SLAM~\citep{yu2025rgbonlygaussiansplattingslam,cheng2025outdoormonocularslamglobal}, and surface reconstruction~\citep{song2024gvkfgaussianvoxelkernel}. Other work uses trajectories for cross-frame alignment, video representations or camera motion generation~\citep{tracktention,trajtok,courant2024exceptional}. These methods provide camera and motion states but typically separate state estimation from language interpretation. Our camera and flow pathways retain both states within the VLM and align them with camera-motion semantics through state--language alignment.

\section{Method}
\label{sec:method}

\subsection{Problem Formulation}
\label{sec:problem}

Answering a camera-motion question requires two things: the motion itself, and the frame in which the question asks to describe it. Neither is given directly. The frame comes from the question rather than the video, and the motion must be inferred from image changes that mix off-screen camera and on-screen scene motion. We therefore formulate camera-motion understanding as reading a dynamic state under a query, and derive from the projection equation what this state must contain.

\paragraph{Camera state and motion semantics.}
To make the camera component explicit, we represent its state at time $t$ as
\begin{equation}
C(t)=
\bigl(
\mathbf{c}(t), R_{WC}(t), K(t)
\bigr),
\label{eq:camera_state}
\end{equation}
where $\mathbf{c}(t)$ is the camera center, $R_{WC}(t)$ its orientation, and
$K(t)$ its intrinsics. Their changes over time describe camera translation,
rotation, and intrinsic changes such as zoom.

\begin{figure}[t]
\centering
\includegraphics[width=\linewidth]{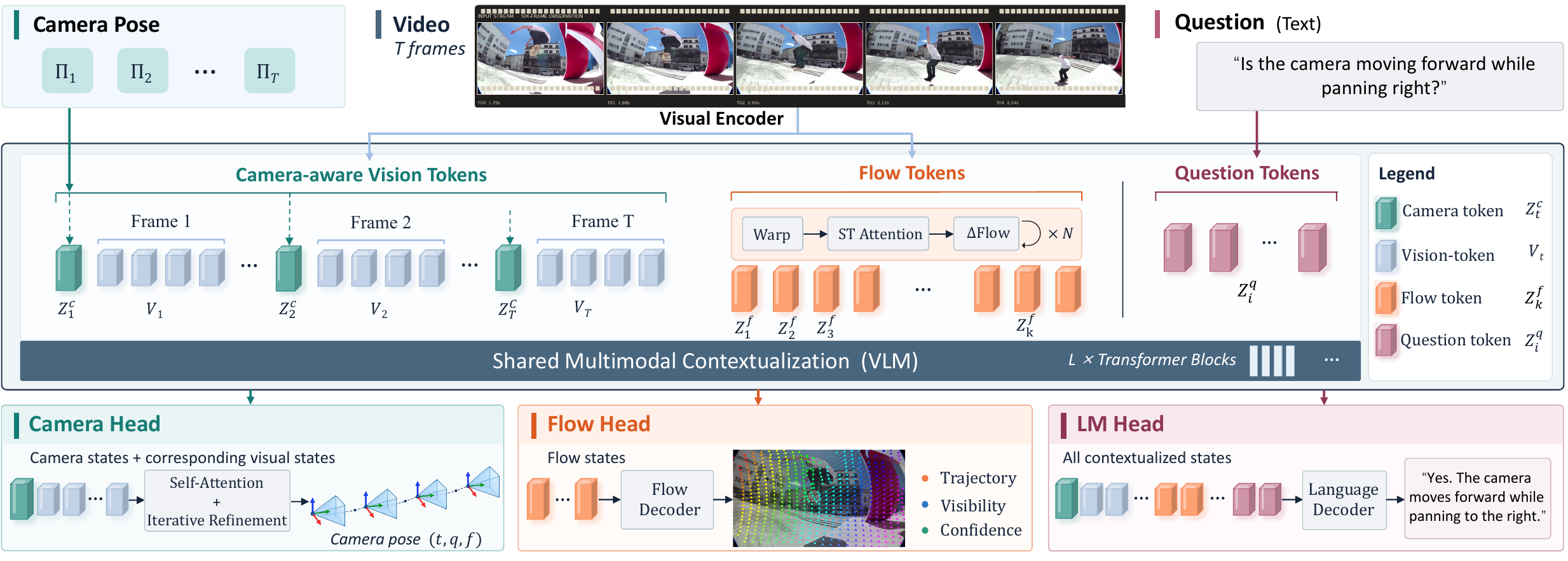}
\vspace{-22pt}
\caption{Architecture of \method{}.
Camera tokens provide continuous camera states, while flow tokens provide
cross-frame motion states. Camera, vision, flow, and question tokens are jointly
contextualized by the shared VLM backbone. The Camera Head predicts continuous
camera parameters, the flow pathway predicts trajectories, visibility, and
confidence, and the LM Head interprets the joint states under the query to
answer camera-motion questions.}
\label{fig:framework}
\vspace{-15pt}
\end{figure}

The resulting dynamic state does not by itself determine its language
description; the interpretation also depends on the relation specified by the
question. We write a query as
\begin{equation}
Q=(a,b,F,\tau),
\label{eq:query}
\end{equation}
where $a$ and $b$ denote the target and reference entities, $F$ the reference
frame, and $\tau$ the temporal interval. The same dynamic state can therefore
support different motion descriptions under different queries. In a follow
shot, for example, the camera moves relative to the scene while remaining
nearly stationary relative to the subject. We formulate camera-motion
understanding as
\begin{equation}
r_Q=\Phi(S_\tau,Q),
\label{eq:motion_relation}
\end{equation}
where $S_\tau$ denotes the continuous dynamic state over the queried interval.

\paragraph{Image-motion ambiguity.}
Image motion does not uniquely identify its physical source. For a scene or
subject point $\mathbf{P}(t)$, its camera-frame position is
\begin{equation}
\mathbf{X}^{C}(t)
=
R_{WC}^{\top}(t)
\left[
\mathbf{P}(t)-\mathbf{c}(t)
\right],
\label{eq:camera_coordinate}
\end{equation}
with image position
$\mathbf{u}(t)=\pi(K(t),\mathbf{X}^{C}(t))$.
Differentiating, as in the image Jacobian of visual
servoing~\citep{chaumette2006visual}, gives
\begin{equation}
\dot{\mathbf{u}}
=
J_XR_{WC}^{\top}
(\dot{\mathbf{P}}-\dot{\mathbf{c}})
-
J_X(\boldsymbol{\omega}_{C}\times\mathbf{X}^{C})
+
J_K\dot K,
\label{eq:image_motion}
\end{equation}
where the three terms capture scene/camera translation, camera rotation, and
intrinsic change. Similar image motion can therefore arise from different
camera and scene dynamics. Pan and lateral translation can produce similar
first-order motion at nearly constant depth, while parallax separates them
across depths. Likewise, zoom and axial translation can produce similar
instantaneous scale changes but differ in their depth dependence. Detailed
derivations are provided in
Appendix~\ref{app:camera_motion_derivations}.

\paragraph{State representation.}
These ambiguities motivate our two complementary state components. Camera states represent camera motion, while flow states represent cross-frame motion and correspondence across subjects and background. We therefore represent the video state as
\begin{equation}
S=
\left(
S_{\mathrm{camera}},
S_{\mathrm{flow}}
\right),
\label{eq:state}
\end{equation}
where $S_{\mathrm{camera}}$ is supervised by camera pose and
$S_{\mathrm{flow}}$ by cross-frame motion. Together, they retain the information
needed to separate different sources of image motion, while
Eq.~\ref{eq:motion_relation} connects the resulting state to the semantics
specified by the query.

\subsection{Architecture}
\label{sec:architecture}

Fig.~\ref{fig:framework} gives an overview of \method{}. Following
Eq.~\ref{eq:state}, we represent the video with two complementary dynamic
states: pose-grounded camera states and trajectory-grounded flow states.
Both are integrated into a shared VLM together with the native video and
question tokens.

\paragraph{VLM backbone.}
We build \method{} on Qwen3.5-VL~\citep{qwen35}. Its native vision encoder
extracts the video features, which are then contextualized by the multimodal
backbone together with camera, flow, and question tokens. The same shared
representation is used for both state prediction and language reasoning, so
pose and flow supervision directly shape the information available to the
language model. No separate visual backbone is introduced for camera or flow
state learning.

The multimodal input is
\begin{equation}
X_{\mathrm{VLM}}
=
\left[
Z_{\mathrm{video+camera}};
Z_{\mathrm{flow}};
Z_Q
\right],
\label{eq:vlm_input}
\end{equation}
where camera tokens are interleaved with the video tokens of their source
frames, $Z_{\mathrm{flow}}$ contains pooled cross-frame motion states, and
$Z_Q$ denotes the question tokens. The shared backbone jointly contextualizes
all token types. The Camera Head predicts camera parameters from the
contextualized camera states, while the LM Head reads the joint representation
to answer the query.

\subsection{Pose-Grounded Camera States}
\label{sec:camera_states}

The camera state should describe how the observer itself moves: its
translation, rotation, and intrinsic change at each sampled frame. These
quantities are not determined by any single frame; they must be inferred from
how the view changes across frames. We therefore form camera states inside the
backbone, where each position can attend to the whole video.

We insert one learnable camera token per source frame next to that frame's
video tokens. After the backbone, the hidden states at these positions form the
camera states $H^c=\{h^c_t\}_{t=1}^{T}$. The Camera Head applies self-attention across the camera states and iteratively refines
per-frame translation, rotation, and field of view,
$\hat Y^c=\mathrm{CameraHead}(H^c)$, which are supervised with camera pose.

Because this supervision acts on $H^c$ rather than on a separate encoder, the
states that the language model reads when answering are the same states that
must predict camera geometry. Camera motion thus remains continuous in the
representation instead of being reduced to discrete labels such as pan, truck,
or zoom, and how these states map to language is determined by the query
(Sec.~\ref{sec:alignment}).

\subsection{Trajectory-Grounded Flow States}
\label{sec:flow_states}

The flow stream complements the camera states with how visible content moves
across frames. As indicated by Eq.~\ref{eq:image_motion}, relative motion and
parallax across subjects, foreground, and background provide important cues for
separating camera motion from scene motion.

Our flow pathway follows CoWTracker~\citep{cowtracker} and operates on features
from the VLM's vision encoder. Point motion is iteratively refined from multi-level visual features through
feature warping and spatio-temporal attention. The resulting point states are decoded into trajectories, visibility,
and confidence and supervised by $\mathcal{L}_{\mathrm{flow}}$.

We initialize points on a uniform image grid so that motion across the full
scene is retained without predefining semantic regions. Their motion states
$\{s_i\}$ are compressed into a fixed set of flow tokens through attention
pooling,
\begin{equation}
Z_{\mathrm{flow}}
=
\mathrm{AttnPool}(\{s_i\}),
\label{eq:flow_tokens}
\end{equation}
which are then passed to the shared VLM backbone. Individual trajectories are
not assigned to camera, subject, or background in advance; their roles are
interpreted jointly with the camera states, visual context, and question.

\subsection{Dynamic-State Learning and Language Alignment}
\label{sec:alignment}

We jointly train the camera and flow states with language supervision so that
they preserve continuous motion information and remain accessible to language
reasoning. The training objective is
\begin{equation}
\mathcal L
=
\mathcal L_{\mathrm{QA}}
+
\lambda_c \mathcal L_{\mathrm{camera}}
+
\lambda_f \mathcal L_{\mathrm{flow}},
\end{equation}
where $\mathcal L_{\mathrm{camera}}$ supervises per-frame translation, rotation,
and field of view, and $\mathcal L_{\mathrm{flow}}$ supervises point
trajectories, visibility, and confidence. The geometric terms keep the states
distinct at a finer granularity than motion labels, while $\mathcal L_{\mathrm{QA}}$
teaches the language model to read them under the entity, reference, and
interval specified by each question.

Training proceeds in two stages. In state-interface adaptation, the language
backbone is frozen, the camera and flow pathways are trained, and the vision
encoder is updated with a small learning rate. The QA loss still backpropagates
through the frozen language model to the new states, so they learn to enter the
representation space the pretrained model already uses, rather than only
fitting pose and flow targets in isolation. In joint state--language
adaptation, the language backbone is also unfrozen and all components are
optimized under the full objective.

Training samples also differ in what the camera tokens receive. In video-only samples, the tokens are learned queries and the camera state must be inferred from the frames. In pose-only samples, ground-truth pose is embedded into the camera tokens without video, so the language model learns how camera geometry maps to camera-motion language. Video+pose samples provide both, tying the inferred state to the explicit geometry. Mixing the three ties camera states from different sources to the same camera-motion language, so that this correspondence enters the shared representation rather than residing only in the camera tokens. At inference, only the video and question are given.

\subsection{Training Data}
\label{sec:training_data}

State--language alignment requires the same clip to carry both kinds of
supervision: pose and flow labels specify what the states should encode, and
QA or captions specify how that content is described.
Using disjoint supervision would weaken the correspondence between learned motion states and their language descriptions.
We therefore pair camera-motion language with pose and flow
labels on the same clips.

Camera-motion language comes from ACaM~\citep{acam} (27K QA) and
CameraBench~\citep{camerabench} (68K QA and captions). Since these videos lack
camera annotations, we estimate camera parameters offline with
VGGT-$\Omega$~\citep{vggtomega} and generate trajectory supervision with
CoWTracker~\citep{cowtracker}. Pairing these signals on the same clips directly
links continuous motion states to their language descriptions.

We further use the video-understanding training data from
Cambrian-P~\citep{cambrianp}, including VSI-590K spatial examples and 510K
general video QA examples from Cambrian-S~\citep{cambrians}, together with the
ground-truth and ViPE-based~\citep{vipe} pseudo camera-pose annotations used by
Cambrian-P. We additionally generate offline CoWTracker trajectory
supervision for a large portion of these videos. Kubric~\citep{kubric} provides
additional ground-truth camera pose and trajectories, and we include 10K spatial
CoT examples from 3DThinker~\citep{thinker3d}. Training and evaluation videos
are strictly disjoint. Detailed sampling ratios are given in
Table~\ref{tab:training_data} (Appendix~\ref{app:exp_data}).

\section{Experiments}

\begin{figure}[!t]
\centering
\includegraphics[width=0.9\linewidth]{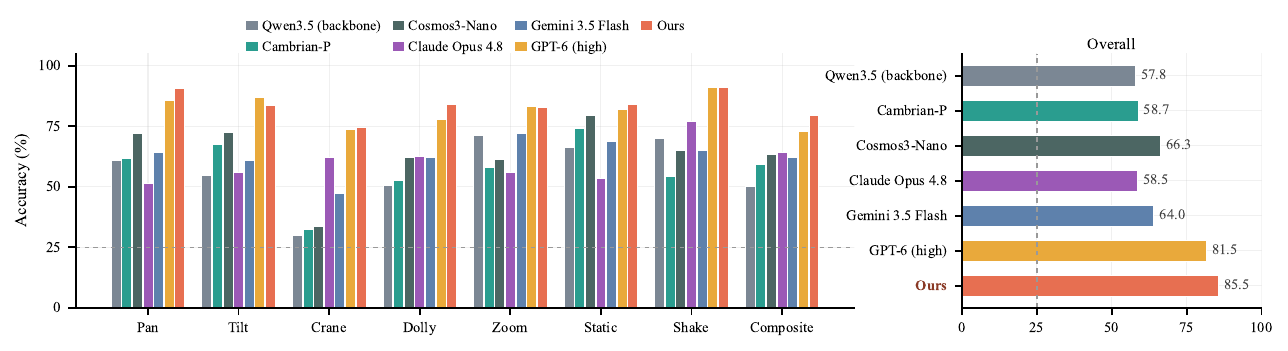}
\vspace{-10pt}
\caption{Per-category accuracy on eight camera-motion primitives, with an
Overall panel. \method{} leads in most categories and is second only to GPT-6
Astra in the others; full numbers for all models are in
Table~\ref{tab:per_category}.}
\label{fig:category}
\vspace{-10pt}
\vspace{16pt}
\captionof{table}{\textbf{Camera-motion understanding accuracy (\%).} Best result per
column in \textbf{bold}, second best \underline{underlined}. \method{} (9B)
surpasses all open-source and camera-motion-specific baselines by a large
margin, exceeds human average accuracy, and is comparable to GPT-6.}
\label{tab:main}
\centering
\scriptsize
\setlength{\tabcolsep}{3pt}
\renewcommand{\arraystretch}{0.88}
\begin{tabular}{llcccc}
\toprule
Model & Params & CameraBench & CineTechBench & ShotBench & Avg \\
\midrule
\multicolumn{6}{l}{\emph{General open-source models}}\\
Qwen2.5-VL \citep{qwen25vl}                & 7B       & 60.00 & 48.67 & 38.36 & 49.01 \\
\rowcolor{black!4}
Qwen3-VL \citep{qwen3vl_tr}                   & 8B       & 62.30 & 37.17 & 40.73 & 46.73 \\
Qwen3.5 (backbone) \citep{qwen35}          & 9B       & 63.64 & 54.87 & 46.98 & 55.16 \\
\rowcolor{black!4}
Qwen3.5 + CoT \citep{qwen35}               & 9B       & 69.20 & 53.98 & 45.47 & 56.22 \\
Orca \citep{orca}                           & 4B       & 62.20 & 52.21 & 36.85 & 50.42 \\
\rowcolor{black!4}
DepthVLM \citep{depthvlm}                  & 4B       & 59.30 & 56.64 & 40.09 & 52.01 \\
Cosmos3-Edge \citep{cosmos3}                & 4B       & 57.60 & 56.64 & 47.41 & 53.88 \\
\rowcolor{black!4}
Cosmos3-Nano \citep{cosmos3}                & 16B      & 70.30 & 67.26 & 57.33 & 64.96 \\
Hy-Embodied-VLM \citep{hyembodiedvlm}      & 30B-A3B  & 65.70 & 53.10 & 47.63 & 55.48 \\
\rowcolor{black!4}
RynnBrain1.1 \citep{rynnbrain}             & 9B       & 65.80 & 56.64 & 43.32 & 55.25 \\
\midrule
\multicolumn{6}{l}{\emph{Geometry-aware models}}\\
Cambrian-P \citep{cambrianp}               & 7B       & 64.30 & 53.09 & 48.06 & 55.15 \\
\rowcolor{black!4}
CamDistill \citep{camdistill}              & 8B       & 40.60 & 42.48 & 43.97 & 42.35 \\
CamInject \citep{camdistill}               & 8B       & 45.50 & 46.90 & 41.81 & 44.74 \\
\rowcolor{black!4}
Spatial-MLLM \citep{spatialmllm}           & --       & 57.70 & 37.17 & 32.11 & 42.33 \\
VLM-3R-7B \citep{vlm3r}                    & 7B       & 57.00 & 33.63 & 39.66 & 43.43 \\
\midrule
\multicolumn{6}{l}{\emph{Camera-motion-specific models}}\\
CameraModel-7B \citep{camerabench}          & 7B       & 70.90 & 52.21 & 44.83 & 55.98 \\
\rowcolor{black!4}
ACaM \citep{acam}                           & 8B       & 63.00 & 57.52 & 56.03 & 58.85 \\
CamReasoner-7B \citep{camreasoner}          & 7B       & 68.00 & 48.67 & 44.83 & 53.83 \\
\rowcolor{black!4}
ShotVL-7B \citep{shotbench}                 & 7B       & 66.10 & 42.48 & 61.21 & 56.60 \\
\midrule
\multicolumn{6}{l}{\emph{Closed-source models}}\\
Claude Opus 4.8                             & --       & 63.49 & 54.87 & 49.03 & 55.80 \\
\rowcolor{black!4}
Gemini 3.5 Flash                            & --       & 66.70 & 61.95 & 58.84 & 62.50 \\
GPT-6 Astra (high)                                & --       & \underline{82.50} & \textbf{83.19} & \underline{79.09} & \underline{81.59} \\
\midrule
Human                                       & --       & 76.20  & 77.61  & 87.30  & 80.37 \\
\midrule
\rowcolor{orange!10}
\method{} (Ours)                            & 9B       & \textbf{88.20} & \underline{77.88} & \textbf{81.47} & \textbf{82.52} \\
\bottomrule
\end{tabular}
\end{figure}

\begin{figure}[!t]
\centering
\includegraphics[width=\linewidth]{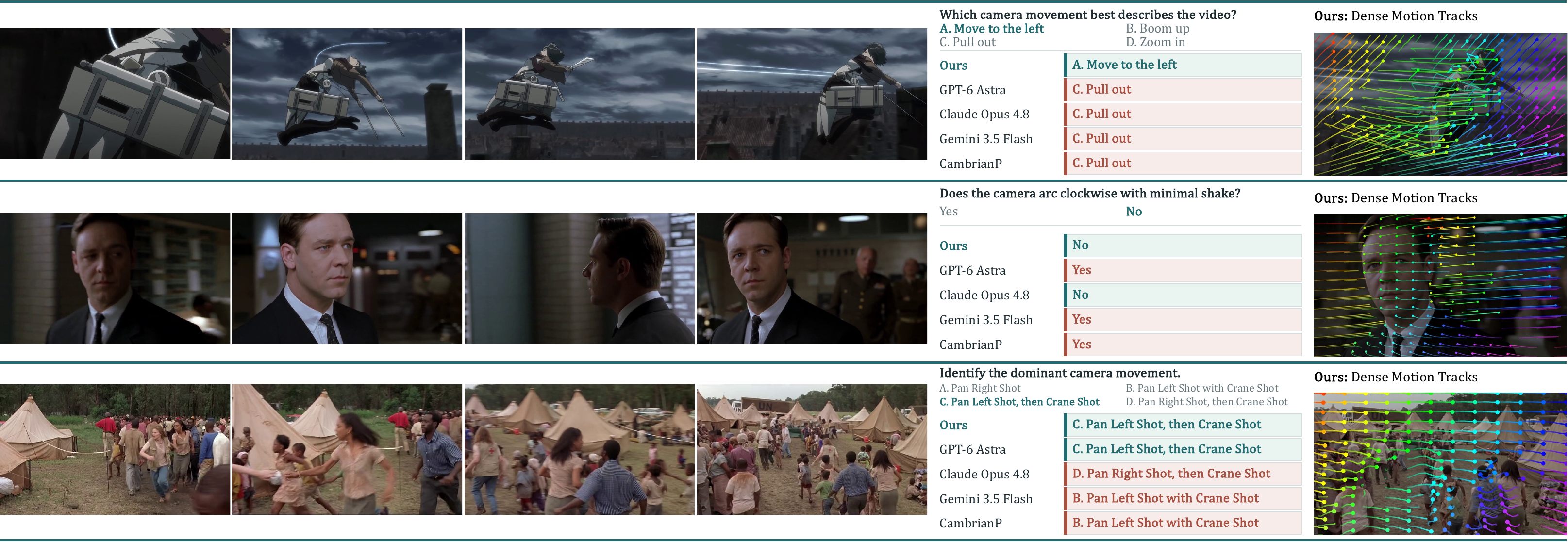}
\vspace{-20pt}
\caption{\textbf{Qualitative comparison.}
Each row compares a model prediction with the bolded ground truth; green denotes
correct predictions and red denotes errors. The rightmost column shows the dense
flow predicted by \method{}. In the first clip, large subject motion causes most
models to mistake a truck-left shot for a
pull-out, while \method{} correctly identifies the lateral camera motion from
background parallax.}
\label{fig:qual}
\vspace{-12pt}
\vspace{12pt}
\captionof{table}{
\textbf{Camera pose estimation.}
ATE, RPE translation, and RPE rotation are reported on ScanNet, TUM RGB-D,
and Sintel; lower is better. The learned camera states retain stable pose information across datasets and
remain competitive with dedicated streaming pose models.
}
\label{tab:pose}
\centering
\scriptsize
\setlength{\tabcolsep}{3pt}
\renewcommand{\arraystretch}{0.88}
\begin{tabular}{lccccccccc}
\toprule
& \multicolumn{3}{c}{ScanNet}
& \multicolumn{3}{c}{TUM RGB-D}
& \multicolumn{3}{c}{Sintel} \\
\cmidrule(lr){2-4}
\cmidrule(lr){5-7}
\cmidrule(lr){8-10}
Model
& ATE & RPE$_t$ & RPE$_r$
& ATE & RPE$_t$ & RPE$_r$
& ATE & RPE$_t$ & RPE$_r$ \\
\midrule

\multicolumn{10}{l}{\emph{Offline}}\\
DUSt3R-GA \citep{dust3r}
& 0.081 & 0.028 & 0.784
& 0.083 & 0.017 & 3.567
& 0.417 & 0.250 & 5.796 \\

MASt3R-GA \citep{mast3r}
& 0.078 & 0.020 & 0.475
& 0.038 & 0.012 & 0.448
& 0.185 & 0.060 & 1.496 \\

MonST3R-GA \citep{monst3r}
& 0.077 & 0.018 & 0.529
& 0.098 & 0.019 & 0.935
& 0.111 & 0.044 & 0.869 \\

Fast3R \citep{fast3r}
& 0.155 & 0.123 & 3.491
& 0.090 & 0.101 & 1.425
& 0.371 & 0.298 & 13.750 \\

FLARE \citep{flare}
& 0.064 & 0.023 & 0.971
& 0.026 & 0.013 & 0.475
& 0.207 & 0.090 & 3.015 \\

VGGT \citep{vggt}
& 0.035 & 0.015 & 0.380
& 0.009 & 0.008 & 0.350
& 0.172 & 0.061 & 0.470 \\

$\pi^3$ \citep{pi3}
& 0.031 & 0.013 & 0.347
& 0.014 & 0.009 & 0.312
& 0.074 & 0.040 & 0.282 \\

MapAnything \citep{mapanything}
& 0.052 & 0.025 & 0.720
& 0.029 & 0.023 & 0.370
& 0.226 & 0.077 & 0.640 \\

\midrule
\multicolumn{10}{l}{\emph{Streaming}}\\
StreamVGGT \citep{streamvggt}
& 0.127 & 0.041 & 1.880
& 0.062 & 0.030 & 0.690
& 0.273 & 0.109 & 0.850 \\

CUT3R \citep{cut3r}
& 0.096 & 0.022 & 0.590
& 0.045 & 0.015 & 0.440
& 0.215 & 0.070 & 0.630 \\

Point3R \citep{point3r}
& 0.097 & 0.035 & 2.791
& 0.058 & 0.031 & 0.758
& 0.442 & 0.154 & 1.897 \\

Spann3R \citep{spann3r}
& 0.096 & 0.023 & 0.661
& 0.056 & 0.021 & 0.591
& 0.329 & 0.110 & 4.471 \\

G2VLM \citep{g2vlm}
& 0.148 & 0.048 & 1.220
& 0.129 & 0.044 & 0.700
& 0.301 & 0.135 & 1.450 \\

Cambrian-P \citep{cambrianp}
& 0.078 & 0.023 & 0.880
& 0.046 & 0.020 & 0.580
& 0.239 & 0.081 & 2.440 \\

\rowcolor{orange!10}
\method{}
& 0.124 & 0.044 & 2.177
& 0.049 & 0.022 & 1.440
& 0.148 & 0.064 & 1.420 \\

\bottomrule
\end{tabular}
\end{figure}

\paragraph{Implementation details.}
We build the model on native Qwen3.5-VL-9B. The visual encoder has $27$
Transformer blocks, the language model has $32$ layers, and the hidden
dimension is $4096$. Videos are uniformly sampled to at most $128$ frames.
Training uses AdamW with a cosine learning-rate schedule: $2\times10^{-6}$ for
the language model and the visual tower, $1\times10^{-4}$ for the camera
module, and $1.5\times10^{-4}$ for the flow module. The first and second
stages run for $20$K and $30$K iterations, respectively, on $32$ H100 GPUs, taking
about $7.5$ days in total. Further
details are given in Appendix~\ref{app:implementation}.

\paragraph{Benchmarks.}
We evaluate camera-motion understanding on CameraBench \citep{camerabench},
CineTechBench \citep{cinetechbench}, and ShotBench \citep{shotbench}.
For CameraBench, we uniformly sample $1{,}000$ Yes/No questions across
camera-motion categories; for CineTechBench and ShotBench, we use all
camera-motion samples. All main evaluations use only the raw video and question,
without external geometric information.
We also evaluate camera pose on TUM RGB-D \citep{tumrgbd},
ScanNet \citep{scannet}, and Sintel \citep{sintel}, and flow on
TAP-Vid DAVIS \citep{tapvid,davis}.

\begingroup
\widowpenalty=10000
\clubpenalty=10000
\paragraph{Human evaluation.}
We further evaluate human performance with $48$ participants, including
professional filmmakers, computer vision students, and camera-motion researchers.
Participants answer randomly assigned Yes/No and four-way multiple-choice
questions.
\par
\endgroup

\subsection{QA evaluation}
\label{sec:qa_eval}

Table~\ref{tab:main} reports the main comparison.
Although CameraModel~\citep{camerabench} and ACaM~\citep{acam} are trained on
camera-motion-specific data, both remain clearly below \method{}.
\method{} consistently outperforms its backbone and all open-source baselines
across the three benchmarks, exceeds average human performance, and remains
competitive with GPT-6. The gain is consistent across benchmarks with different motion distributions, a property that single-benchmark fine-tuning does not exhibit.

Figure~\ref{fig:category} reports performance across eight camera-motion
categories. \method{} performs consistently well across all categories and
achieves the best overall accuracy, showing that the gains are broadly
distributed across different motion types. Figure~\ref{fig:qual} presents qualitative comparisons on representative clips.
In the first example, substantial subject motion causes most models to mistake
a truck-left shot for a pull-out, while \method{} correctly recovers the camera
motion using the background motion cues. The predicted dense flow further
visualizes these cross-frame dynamics, helping disambiguate camera and subject motion. Additional examples are provided in
Appendix~\ref{app:more_qual}, with full videos in the supplementary material.

\begin{table}[!t]
\centering
\begin{minipage}[t]{0.53\linewidth}
\centering
\caption{
\textbf{Point tracking on TAP-Vid DAVIS.}
AJ, $\delta_{\mathrm{avg}}$, and OA are reported; higher is better. The learned flow states retain
cross-frame tracking information after integration into the VLM.
}
\label{tab:flow}

\scriptsize
\setlength{\tabcolsep}{3pt}
\renewcommand{\arraystretch}{0.88}
\resizebox{\linewidth}{!}{\begin{tabular}{llccc}
\toprule
Model & Train data & AJ & $\delta_{\mathrm{avg}}$ & OA \\
\midrule
TAPIR \citep{tapir}              & Kub       & 56.2 & 70.0 & 86.5 \\
CoTracker \citep{cotracker}      & Kub       & 61.8 & 76.1 & 88.3 \\
TAPTR \citep{taptr}              & Kub       & 63.0 & 76.1 & 91.1 \\
LocoTrack \citep{locotrack}      & Kub       & 62.9 & 75.3 & 87.2 \\
BootsTAPIR \citep{bootstapir}    & Kub+15M   & 61.4 & 73.6 & 88.7 \\
CoTracker3 \citep{cotracker3} (online)  & Kub+15k & 63.8 & 76.3 & 90.2 \\
CoTracker3 \citep{cotracker3} (offline) & Kub+15k & 64.4 & 76.9 & 91.2 \\
DOT \citep{dot_tracking}         & Kub       & 60.1 & 74.5 & 89.0 \\
DELTA \citep{delta_track}        & Kub       & 60.8 & 75.2 & 87.6 \\
AllTracker \citep{alltracker} & Kub     & 61.9 & 75.4 & 87.8 \\
AllTracker \citep{alltracker} & Kub+Mix & 63.3 & 76.3 & 90.1 \\
CoWTracker \citep{cowtracker} & Kub     & 65.5 & 78.0 & 92.1 \\
\midrule
\rowcolor{orange!10}
Ours               & Kub+CoW distill        & 50.0 & 63.2 & 88.2 \\
\bottomrule
\end{tabular}}
\end{minipage}\hfill
\begin{minipage}[t]{0.44\linewidth}
\centering
\caption{
Ablation study of the main components and training strategy. Two-stage
variants are trained for $10$K iterations per stage; one-stage training runs only
the $10$K-iteration state-interface stage. Numbers are therefore lower than
in Table~\ref{tab:main}.
}
\label{tab:main_ablation}
\scriptsize
\setlength{\tabcolsep}{3pt}
\renewcommand{\arraystretch}{0.88}
\resizebox{\linewidth}{!}{\begin{tabular}{lcccc}
\toprule
Method
& CameraBench
& CineTech
& ShotBench
& Avg. \\
\midrule
Backbone
& 63.64 & 54.87 & 46.98 & 55.2 \\

w/o Flow
& 64.50 & 55.75 & 72.20 & 64.1 \\

w/o Camera \& Flow
& 67.20 & 57.10 & 42.10 & 55.5 \\

One-stage Training
& 64.50 & 64.00 & 53.88 & 60.8 \\

\textbf{Full Model}
& \textbf{68.20} & \textbf{67.88} & \textbf{70.79} & \textbf{69.0} \\
\bottomrule
\end{tabular}}
\end{minipage}
\end{table}

\subsection{State Evaluation}
\label{sec:state_eval}

We further evaluate the learned camera and flow states to verify that they retain the continuous motion information introduced by their corresponding supervision. This provides a direct check that the model preserves explicit motion states beyond the final language predictions.

\paragraph{Camera state.}
We evaluate pose estimation from the camera states (Table~\ref{tab:pose}) to verify that they
retain meaningful camera-motion information. Baseline results follow \citet{cambrianp}. Although several baselines are trained on substantially broader geometric data, our model still achieves competitive pose estimation. On Sintel, it obtains the best ATE and RPE$_t$ among all streaming baselines, showing that the camera states learn and preserve transferable camera-motion information.

\paragraph{Flow state.}
We evaluate point tracking from the flow states on TAP-Vid DAVIS (Table~\ref{tab:flow}). Baseline results follow \citet{cowtracker}.
Although the flow pathway is integrated into the VLM for camera-motion
understanding rather than optimized as a standalone tracker, it retains
meaningful cross-frame tracking information, including competitive occlusion
accuracy. This confirms that the learned flow states preserve correspondence
and motion information after joint training.

\subsection{Ablation Study}
\label{sec:ablation}

Table~\ref{tab:main_ablation} evaluates the contributions of the state
representation and supervision. For a controlled comparison, two-stage
variants are trained for $10$K iterations per stage, and one-stage training runs
only the $10$K-iteration state-interface stage without joint state--language
adaptation; the full model here therefore differs from the main result in
Table~\ref{tab:main}.
Removing flow states degrades performance,
showing that cross-frame motion complements the camera states. Without camera
and flow supervision, performance drops close to the backbone, confirming that
explicit state supervision is important for learning useful dynamic
representations. The full training strategy further improves performance over
one-stage training, indicating that state learning and state--language alignment are complementary.

The Appendix further shows that inference-time masking causes only limited
degradation (Table~\ref{tab:token_masking}), suggesting that dynamic information
is distributed across the shared representation, and that the model is robust
to most flow configurations (Table~\ref{tab:hyperparameter}).

\begin{samepage}
\subsection{Limitations}
\label{sec:limitations}

Some training videos rely on offline supervision generated by
VGGT-$\Omega$ and CoWTracker, so state learning may inherit errors from
upstream estimation. Rare compound motions and unusual imaging conditions are
also less represented in the training data. Finally, our current setting
focuses on shot-level camera motion and does not yet model camera-motion
relations across cuts or longer cinematographic sequences.

\end{samepage}

\section{Conclusion}

We represent camera motion through continuous camera and flow states aligned
with camera-motion semantics. Keeping both off-screen camera motion and
on-screen image motion in the representation yields consistent gains across
camera-motion benchmarks, particularly on spatial direction and compound camera
motion. This provides a basis for interpreting shot composition and
subject--camera relations, and for more controllable video generation and
editing.




\bibliography{refs_arxiv}
\bibliographystyle{camera_refs}

\clearpage
\appendix
\section{Derivations for Camera-Motion Formulation}
\label{app:camera_motion_derivations}

\subsection{Relative Motion under a Moving Reference Frame}
\label{app:relative_motion}

\paragraph{Reference-conditioned relative motion.}
For a target entity $a$, reference entity $b$, and reference frame $F$, their
relative displacement is
\begin{equation}
\mathbf{d}^{F}_{a|b}(t)
=
R_{WF}^{\top}(t)
\left[
\mathbf{x}_a^W(t)-\mathbf{x}_b^W(t)
\right].
\label{eq:relative_displacement}
\end{equation}
Let
\begin{equation}
\mathbf{y}
=
\mathbf{x}_a^W-\mathbf{x}_b^W.
\end{equation}
If $\boldsymbol{\omega}_F$ is the angular velocity of $F$ relative to $W$,
expressed in $F$, then
\begin{equation}
\frac{d}{dt}R_{WF}^{\top}
=
-[\boldsymbol{\omega}_F]_{\times}R_{WF}^{\top}.
\end{equation}
Hence,
\begin{align}
\frac{d}{dt}\mathbf{d}_{a|b}^{F}
&=
\frac{dR_{WF}^{\top}}{dt}\mathbf{y}
+
R_{WF}^{\top}\dot{\mathbf{y}}
\\
&=
R_{WF}^{\top}
\left(
\dot{\mathbf{x}}_a^W-\dot{\mathbf{x}}_b^W
\right)
-
\boldsymbol{\omega}_F
\times
\mathbf{d}_{a|b}^{F},
\label{eq:relative_velocity}
\end{align}
which is the relative velocity expressed in the moving frame $F$.

For a static relation, the world-frame positions remain fixed while the
coordinate representation depends on $F$. If two reference frames have
opposite horizontal axes, the same displacement has opposite horizontal signs.

For a follow shot, let $a$ be the subject and $b$ the camera. When
\begin{equation}
\dot{\mathbf{x}}_a^W
\approx
\dot{\mathbf{x}}_b^W
\end{equation}
and reference-frame rotation is small over the interval,
$\mathbf{v}_{a|b}^{C}\approx 0$, while both entities may have nonzero
world-frame velocity.

\subsection{Camera-Frame Motion Decomposition}
\label{app:camera_frame}

For a scene point $\mathbf{P}(t)$ with camera-frame position
$\mathbf{X}^{C}=R_{WC}^{\top}(\mathbf{P}-\mathbf{c})$, differentiating gives
\begin{equation}
\dot{\mathbf{X}}^{C}
=
R_{WC}^{\top}
(\dot{\mathbf{P}}-\dot{\mathbf{c}})
-
\boldsymbol{\omega}_{C}\times\mathbf{X}^{C},
\label{eq:camera_decomposition}
\end{equation}
where the terms correspond to scene/subject motion minus camera translation,
and camera rotation, respectively.

\subsection{Pan--Truck Ambiguity}
\label{app:pan_truck}

Consider a static scene point
$\mathbf{X}^C=(X,Y,Z)^\top$ with normalized image coordinates
\begin{equation}
x=\frac{X}{Z},
\qquad
y=\frac{Y}{Z}.
\end{equation}

For pure yaw
$\boldsymbol{\omega}_{C}=(0,\omega_y,0)^\top$ with no translation,
Eq.~\ref{eq:camera_decomposition} gives
\begin{equation}
\dot{\mathbf{X}}^C
=
-\boldsymbol{\omega}_{C}
\times
\mathbf{X}^C.
\end{equation}
Therefore,
\begin{equation}
\dot X=-\omega_y Z,
\qquad
\dot Y=0,
\qquad
\dot Z=\omega_y X,
\end{equation}
and
\begin{equation}
\dot x_{\mathrm{pan}}
=
-\omega_y(1+x^2),
\qquad
\dot y_{\mathrm{pan}}
=
-\omega_yxy.
\label{eq:pan_flow}
\end{equation}

For pure lateral camera translation $T_x$,
\begin{equation}
\dot X=-T_x,
\qquad
\dot Y=0,
\qquad
\dot Z=0,
\end{equation}
which gives
\begin{equation}
\dot x_{\mathrm{truck}}
=
-\frac{T_x}{Z},
\qquad
\dot y_{\mathrm{truck}}
=
0.
\label{eq:truck_flow}
\end{equation}

Let
\begin{equation}
Z=Z_0+\delta Z,
\qquad
|x|,|y|\leq\epsilon,
\qquad
\frac{T_x}{Z_0}=\omega_y.
\end{equation}
Using
\begin{equation}
\frac{Z_0}{Z_0+\delta Z}
=
1-\frac{\delta Z}{Z_0}
+
O\!\left(
\frac{\delta Z^2}{Z_0^2}
\right),
\end{equation}
we obtain
\begin{align}
\dot x_{\mathrm{pan}}
-
\dot x_{\mathrm{truck}}
&=
O(\epsilon^2)
+
O\!\left(
\frac{\delta Z}{Z_0}
\right),
\\
\dot y_{\mathrm{pan}}
-
\dot y_{\mathrm{truck}}
&=
O(\epsilon^2).
\end{align}
The first-order ambiguity disappears through depth-dependent parallax,
higher-order perspective terms, or direct camera-pose information.

\subsection{Zoom--Dolly Ambiguity}
\label{app:zoom_dolly}

Let $\mathbf{p}$ denote image coordinates relative to the principal point:
\begin{equation}
\mathbf{p}
=
f
\begin{bmatrix}
X/Z\\
Y/Z
\end{bmatrix}.
\end{equation}

For optical zoom with fixed camera pose,
\begin{equation}
\dot{\mathbf{p}}_{\mathrm{zoom}}
=
\frac{\dot f}{f}
\mathbf{p}.
\label{eq:zoom_flow}
\end{equation}

For forward axial camera translation with velocity $T_z$,
\begin{equation}
\dot Z=-T_z,
\end{equation}
which gives
\begin{equation}
\dot{\mathbf{p}}_{\mathrm{dolly}}
=
\frac{T_z}{Z}
\mathbf{p}.
\label{eq:dolly_flow}
\end{equation}

If all visible points lie at depth $Z_0$ and
\begin{equation}
\frac{\dot f}{f}
=
\frac{T_z}{Z_0},
\end{equation}
the two instantaneous radial flows coincide.

For two visible points with $Z_i\neq Z_j$, dolly produces radial rates
\begin{equation}
\frac{T_z}{Z_i}
\qquad\text{and}\qquad
\frac{T_z}{Z_j},
\end{equation}
while zoom produces the same $\dot f/f$ for both. The spatial variation of the
flow therefore separates axial translation from focal-length change.

\subsection{Reflection}
\label{app:reflection}

For a planar mirror with unit normal $\mathbf n$, the reflection matrix is
\begin{equation}
M
=
I-2\mathbf n\mathbf n^{\top},
\qquad
\det(M)=-1.
\end{equation}
A motion vector transforms as
\begin{equation}
\mathbf v_{\mathrm{refl}}
=
M\mathbf v.
\end{equation}
Its normal component changes sign,
\begin{equation}
\mathbf n^{\top}\mathbf v_{\mathrm{refl}}
=
-\mathbf n^{\top}\mathbf v,
\end{equation}
while tangential components are unchanged. The resulting image direction
depends on the mirror orientation and the queried reference frame. A
left--right reversal occurs when this sign change projects onto the horizontal
axis of that frame.

\subsection{Query-Induced State Equivalence}
\label{app:query_equivalence}

Let $\mathcal{Q}_{\mathrm{tr}}$ denote the training query set. Define
\begin{equation}
S_1
\sim_{\mathcal{Q}_{\mathrm{tr}}}
S_2
\quad\Longleftrightarrow\quad
\Phi(S_1,Q)
=
\Phi(S_2,Q),
\qquad
\forall Q\in\mathcal{Q}_{\mathrm{tr}}.
\end{equation}

For
\begin{equation}
\mathcal{L}_{\mathrm{QA}}
=
-\log p(A\mid V,Q),
\end{equation}
states in the same equivalence class produce the same target answers for all
observed queries. Mapping them to the same internal representation incurs no
additional QA loss.

There may still exist
$Q^\star\notin\mathcal{Q}_{\mathrm{tr}}$ such that
\begin{equation}
\Phi(S_1,Q^\star)
\neq
\Phi(S_2,Q^\star).
\end{equation}
Answer equivalence under the training queries therefore does not imply state
equivalence.

Camera-pose supervision constrains camera translation, rotation, and viewpoint
change. Scene-wide flow supervision constrains cross-frame motion and
correspondence across image regions. Without metric depth supervision,
world-frame object translation remains ambiguous in scale.

\section{Implementation Details}
\label{app:implementation}

\method{} builds on Qwen3.5-VL-9B. Videos are uniformly sampled to at most
$128$ frames ($64$ temporal groups). The vision tower ($27$ blocks) keeps its
native spatial merger and multimodal RoPE; the language backbone has $32$
layers with $d=4096$.

\paragraph{Camera pathway.}
Each source frame carries one camera token. Two learnable query embeddings
$q_{\mathrm{first}}$ and $q_{\mathrm{shared}}$ parameterize the token
positions; frame-specific states emerge from multimodal contextualization.
Camera hidden states are projected to $2048$-D and consumed by a VGGT-style
camera head~\citep{vggt,cambrianp} (four self-attention layers, four
refinement iterations) that predicts translation, quaternion rotation and
field of view per frame.

\paragraph{Flow pathway.}
Unmerged visual features from four intermediate layers are fused through a DPT
decoder ($128$ channels) and fed to the CoWTracker tracking head (three refinement iterations) to predict
trajectories, visibility and confidence on a fixed $16\times16$ grid ($256$
anchor points). Learned-query attention pooling ($64$ queries) compresses
these anchor states into $64$ flow tokens in the $4096$-D language space. Each
pooled token is residually anchored to the mean video embedding.

\paragraph{Sequence layout.}
Camera tokens are interleaved with their source frames; flow tokens are
inserted before the question:
$X_{\mathrm{LM}} = [Z_{\mathrm{video{+}camera}};\,Z_{\mathrm{flow}};\,Z_Q]$.
All state tokens are computed in prefill and cached; they occupy multimodal
input positions and are excluded from vocabulary generation.

\paragraph{Supervision.}
Flow: pixel-space Huber loss on trajectories plus binary cross-entropy on
visibility and confidence. Camera: multi-stage regression on translation,
rotation and field of view. CoWTracker produces offline pseudo labels and does not
participate in inference.

\paragraph{Optimization.}
AdamW with cosine schedule. Learning rates: $2{\times}10^{-6}$ (LM and vision
tower), $1{\times}10^{-5}$ (merger and general modules), $1{\times}10^{-4}$
(camera module), $1.5{\times}10^{-4}$ (flow modules). The two stages run
$20$K and $30$K iterations at batch size $256$; total training takes about
$7.5$ days.

\paragraph{Training data.}
\label{app:exp_data}
Table~\ref{tab:training_data} lists the training sources together with their
language, camera, and flow supervision and their sampling ratios.

\begin{table}[!ht]
\centering
\caption{
\textbf{Training data composition.}
Sampling ratios are normalized across all training sources.
Camera supervision is provided by ground-truth poses or offline pseudo poses,
while flow supervision is provided by ground-truth trajectories or
CoWTracker teacher labels.
}
\label{tab:training_data}
\scriptsize
\setlength{\tabcolsep}{3.5pt}
\renewcommand{\arraystretch}{0.95}
\begin{tabular}{lcccr}
\toprule
Data source
& Language
& Camera supervision
& Flow supervision
& Ratio \\
\midrule

ACaM~\citep{acam}
& QA
& VGGT-$\Omega$
& CoWTracker
& 15.05\% \\

CameraBench VQA~\citep{camerabench}
& QA
& VGGT-$\Omega$
& CoWTracker
& 15.05\% \\

CameraBench Caption~\citep{camerabench}
& Caption
& VGGT-$\Omega$
& CoWTracker
& 3.23\% \\

Cambrian-P / Cambrian-S~\citep{cambrianp,cambrians}
& Spatial + Video QA
& GT / ViPE pseudo pose
& CoWTracker
& 48.39\% \\

Kubric~\citep{kubric}
& --
& GT
& GT
& 10.75\% \\

3DThinker~\citep{thinker3d}
& CoT
& --
& CoWTracker
& 7.53\% \\

\bottomrule
\end{tabular}

\vspace{2pt}

\end{table}

\FloatBarrier
\section{Additional Experiments}
\label{app:additional_exp}

\subsection{Inference-time State Token Masking}
\label{app:token_masking}

\begin{table}[!ht]
\centering
\caption{
Inference-time masking of state tokens.
All variants use the fully trained model and differ only in the state tokens
masked during inference.
}
\label{tab:token_masking}
\scriptsize
\setlength{\tabcolsep}{3pt}
\renewcommand{\arraystretch}{0.88}
\begin{tabular}{lcccc}
\toprule
Inference Setting
& CameraBench
& CineTech
& ShotBench
& Avg.\ \\
\midrule
Full
& 88.20 & 77.88 & 81.47 & 82.5 \\
Mask Flow Tokens
& 88.51 & 75.30 & 81.04 & 81.6 \\
Mask Camera Tokens
& 88.30 & 76.19 & 81.92 & 82.1 \\
Mask Both
& 88.81 & 77.07 & 81.71 & 82.5 \\
\bottomrule
\end{tabular}
\end{table}

Masking camera, flow, or both state-token groups at inference produces only
limited performance changes, indicating that the learned dynamic information
is distributed across the shared multimodal representation after
state--language alignment.

\FloatBarrier
\subsection{Hyperparameter Sensitivity}
\label{app:hyperparameter}

Table~\ref{tab:hyperparameter} shows that the model is largely insensitive to
the configuration of the flow pathway. Varying the number of flow tokens,
spatial grid, temporal window, or refinement iterations changes CineTech
accuracy by at most 1.69 points. In contrast, reducing the input to 64 and
32 frames decreases accuracy by 4.35 and 7.01 points, respectively, indicating
that sufficient temporal coverage is important for camera-motion understanding.

\begin{table}[!ht]
\centering
\caption{
Hyperparameter sensitivity analysis.
Only the specified configuration is changed while all other settings remain fixed.
}
\label{tab:hyperparameter}
\scriptsize
\setlength{\tabcolsep}{3pt}
\renewcommand{\arraystretch}{0.88}
\begin{tabular}{llc}
\toprule
Setting & Value & CineTech \\
\midrule
Default
& -- & \textbf{77.88} \\
\midrule
Flow Tokens
& 16  & 77.07 \\
& 32  & 77.07 \\
& 128 & 77.07 \\
\midrule
Grid
& 8  & 77.07 \\
& 32 & 77.07 \\
\midrule
Window
& 8  & 76.19 \\
& 32 & 77.07 \\
\midrule
Iterations
& 1 & 76.19 \\
& 5 & 77.07 \\
\midrule
Frames
& 64 & 73.53 \\
& 32 & 70.87 \\
\bottomrule
\end{tabular}
\end{table}

\FloatBarrier
\begin{samepage}
\section{Additional Results}
\label{app:additional_results}

\subsection{Full Per-Category Results}
\label{app:per_category}

Table~\ref{tab:per_category} reports accuracy on every camera-motion primitive
category for representative models from Table~\ref{tab:main}. Overall is the accuracy pooled over every
evaluated item across the three benchmarks, so it need not equal the
unweighted Avg.\ column of Table~\ref{tab:main}. \method{} achieves the
highest overall accuracy ($85.48\%$) and leads in most individual categories,
with particularly strong gains on geometrically challenging primitives such as
Crane, Dolly and Truck, where many baselines fall below $55\%$.

\end{samepage}

\begin{table}[!ht]
\caption{\textbf{Per-category camera-motion accuracy (\%).} Overall pools
every evaluated item across CameraBench, CineTechBench and ShotBench.}
\label{tab:per_category}
\centering
\tiny
\renewcommand{\arraystretch}{0.92}
\setlength{\tabcolsep}{2.3pt}
\begin{tabular}{lccccccccccccccc}
\toprule
Model & Overall & Arc & Composite & Crane & Dolly & Other & Pan & Pedestal & Roll & Shake & Static & Tilt & Tracking & Truck & Zoom \\
\midrule
\multicolumn{16}{l}{\emph{Open-source models}}\\
ACaM                   & 60.56 & 50.00 & 56.12 & 55.41 & 51.65 & 61.65 & 53.12 & 90.00 & 69.09 & 78.95 & 70.71 & 52.41 & 100.00 & 56.74 & 74.42 \\
\rowcolor{black!4}
Cambrian-P             & 58.72 & 40.00 & 59.18 & 32.43 & 52.48 & 63.91 & 61.61 & 60.00 & 65.45 & 54.39 & 74.06 & 67.59 & 60.00 & 40.43 & 58.14 \\
CamDistill             & 41.72 & 40.00 & 36.73 & 33.78 & 37.60 & 35.34 & 46.43 & 40.00 & 45.45 & 45.61 & 49.79 & 50.34 & 60.00 & 41.13 & 27.91 \\
\rowcolor{black!4}
CamInject              & 44.51 & 55.00 & 39.80 & 43.24 & 38.43 & 47.37 & 42.41 & 40.00 & 56.36 & 47.37 & 55.65 & 42.07 & 60.00 & 47.52 & 31.01 \\
Cosmos3-Edge           & 54.53 & 20.00 & 56.12 & 25.68 & 40.08 & 56.39 & 60.71 & 80.00 & 61.82 & 52.63 & 63.18 & 63.45 & 60.00 & 48.94 & 65.12 \\
\rowcolor{black!4}
Cosmos3-Nano           & 66.27 & 75.00 & 63.27 & 33.78 & 61.98 & 75.94 & 71.88 & 70.00 & 61.82 & 64.91 & 79.50 & 72.41 & 70.00 & 51.06 & 61.24 \\
DepthVLM               & 53.46 & 40.00 & 42.86 & 29.73 & 38.84 & 66.17 & 57.59 & 90.00 & 49.09 & 61.40 & 68.62 & 53.10 & 70.00 & 43.97 & 61.24 \\
\rowcolor{black!4}
Hy-Embodied-VLM-1.0    & 59.48 & 60.00 & 54.08 & 31.08 & 53.72 & 69.92 & 59.38 & 80.00 & 70.91 & 64.91 & 75.31 & 53.79 & 60.00 & 46.81 & 62.02 \\
Orca                   & 54.03 & 40.00 & 55.10 & 27.03 & 45.04 & 62.41 & 50.00 & 70.00 & 61.82 & 54.39 & 71.97 & 56.55 & 90.00 & 36.88 & 61.24 \\
\rowcolor{black!4}
Qwen3.5 (our backbone) & 57.77 & 30.00 & 50.00 & 29.73 & 50.41 & 64.66 & 60.71 & 80.00 & 63.64 & 70.18 & 66.11 & 54.48 & 70.00 & 50.35 & 71.32 \\
Qwen3.5 + CoT          & 61.13 & 45.00 & 45.92 & 43.24 & 54.55 & 70.68 & 64.29 & 60.00 & 52.73 & 84.21 & 71.55 & 62.07 & 60.00 & 49.65 & 68.22 \\
\rowcolor{black!4}
RynnBrain1.1           & 58.53 & 50.00 & 46.94 & 48.65 & 61.57 & 69.92 & 61.61 & 40.00 & 67.27 & 59.65 & 55.65 & 64.14 & 60.00 & 41.84 & 65.89 \\
\midrule
\multicolumn{16}{l}{\emph{Closed-source models}}\\
Claude Opus 4.8        & 58.47 & 50.00 & 64.29 & 62.16 & 62.40 & 62.41 & 51.34 & 90.00 & 61.82 & 77.19 & 53.56 & 55.86 & 70.00 & 56.03 & 55.81 \\
\rowcolor{black!4}
Gemini 3.5 Flash       & 64.05 & 50.00 & 62.24 & 47.30 & 61.98 & 73.68 & 64.29 & 60.00 & 60.00 & 64.91 & 68.62 & 60.69 & 70.00 & 59.57 & 72.09 \\
GPT-6 (high)           & 81.55 & 75.00 & 72.73 & 73.81 & 77.88 & 79.33 & 85.80 & 90.00 & 83.64 & 91.23 & 82.02 & 86.79 & 100.00 & 80.49 & 83.05 \\
\midrule
Human                  & 73.16 & 50.00 & 61.70 & 40.00 & 71.60 & 55.56 & 70.42 & 80.00 & 89.47 & 66.67 & 87.32 & 80.95 & 90.91 & 80.00 & 82.14 \\
\midrule
\rowcolor{orange!10}
Ours & 85.48 & 100.00 & 79.59 & 74.32 & 83.88 & 90.23 & 90.63 & 80.00 & 87.27 & 91.23 & 84.10 & 83.45 & 80.00 & 87.94 & 82.95 \\
\bottomrule
\end{tabular}
\end{table}

\FloatBarrier
\subsection{Qualitative Examples}
\label{app:more_qual}

\begin{figure}[!ht]
\centering
\includegraphics[width=\linewidth]{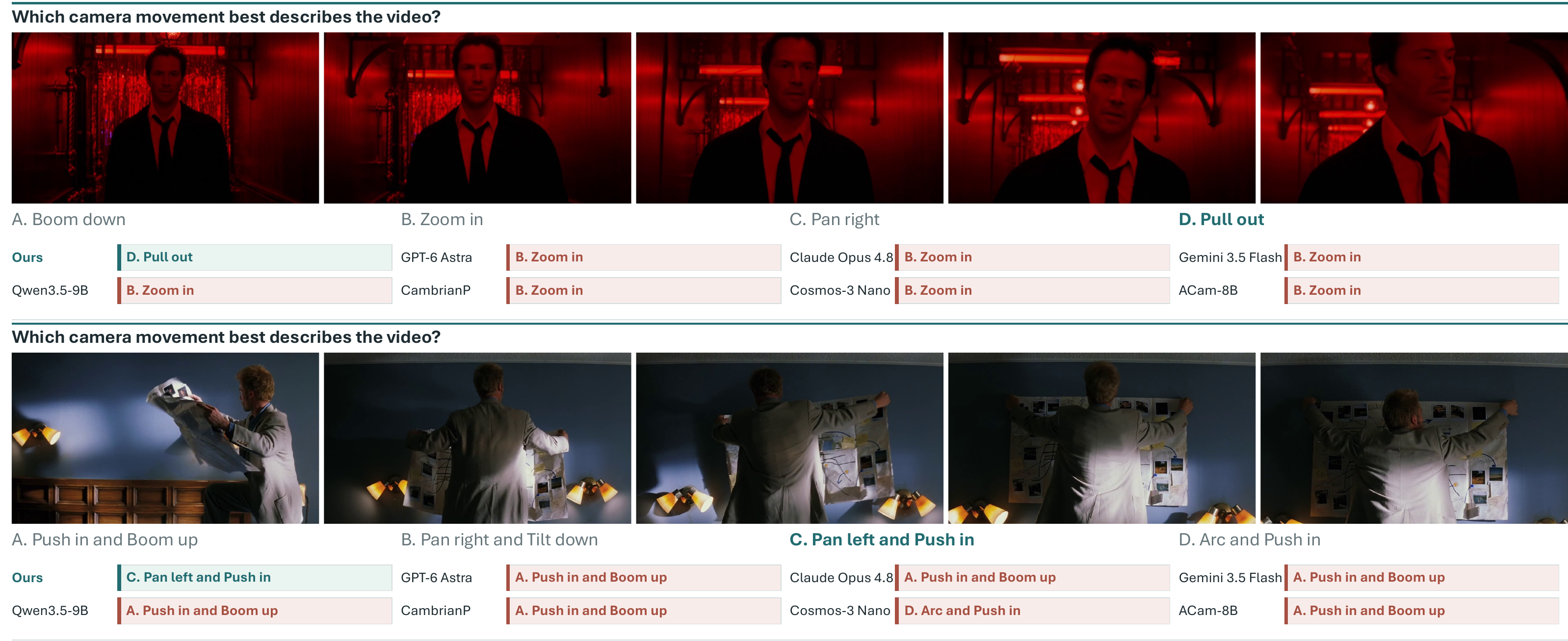}
\caption{Qualitative examples on pull-out and compound pan-left-push-in motions. Most baselines confuse pull-out with zoom-in; only \method{} correctly identifies both clips.}
\label{fig:suppqa1}
\end{figure}

\begin{figure}[!ht]
\centering
\includegraphics[width=\linewidth]{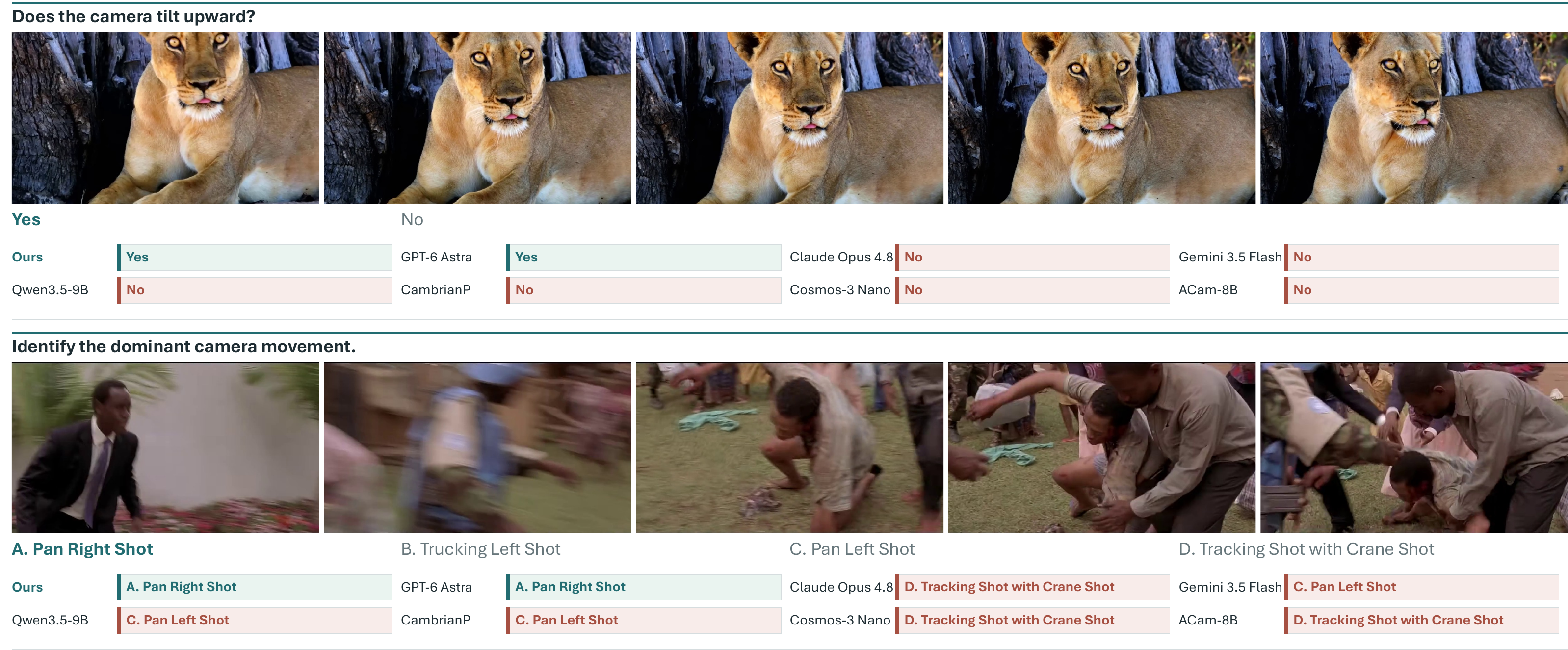}
\caption{Qualitative examples on tilt-up and pan-right recognition. Most baselines fail on the binary tilt question; several confuse pan direction or hallucinate a crane motion.}
\label{fig:suppqa2}
\end{figure}

\begin{figure}[!ht]
\centering
\includegraphics[width=\linewidth]{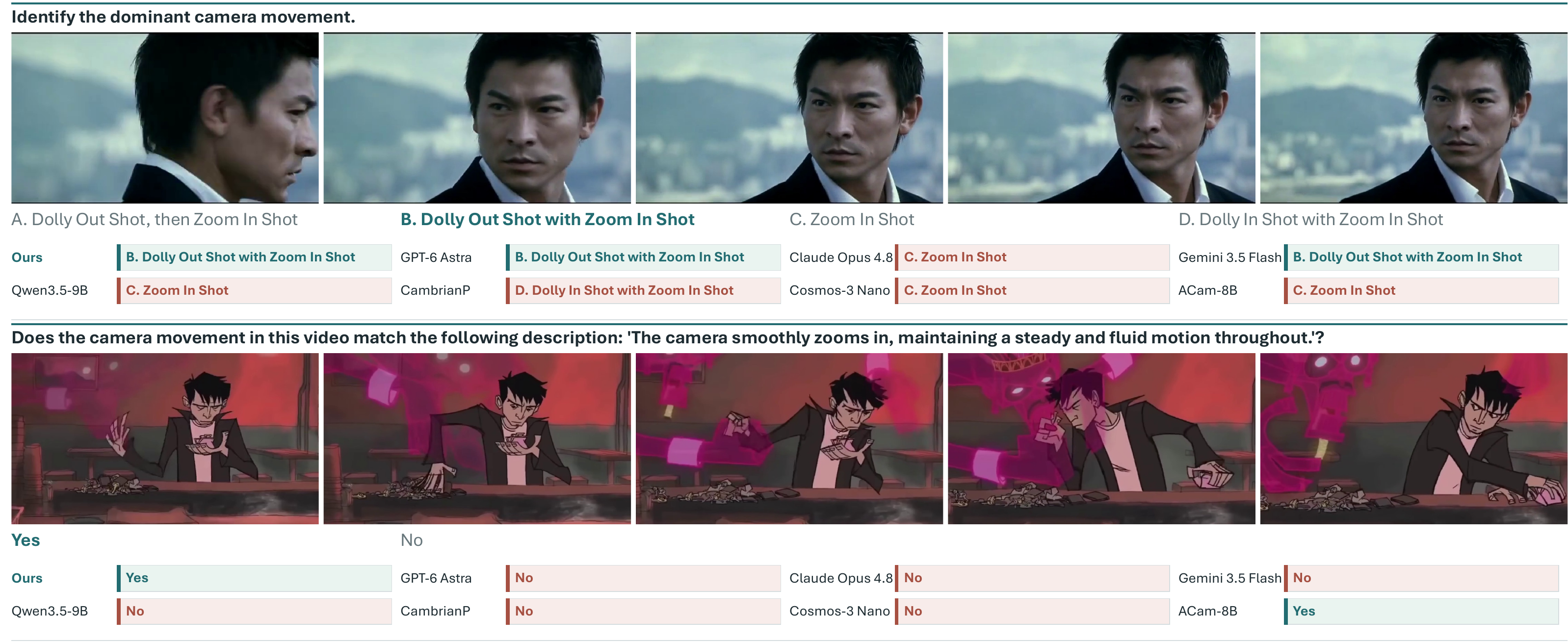}
\caption{Qualitative examples on compound dolly-zoom and zoom-in description matching. The dolly-out-with-zoom-in clip tests fine-grained compound-motion parsing; the description-matching clip requires verifying motion smoothness.}
\label{fig:suppqa3}
\end{figure}

\begin{figure}[!ht]
\centering
\includegraphics[width=\linewidth]{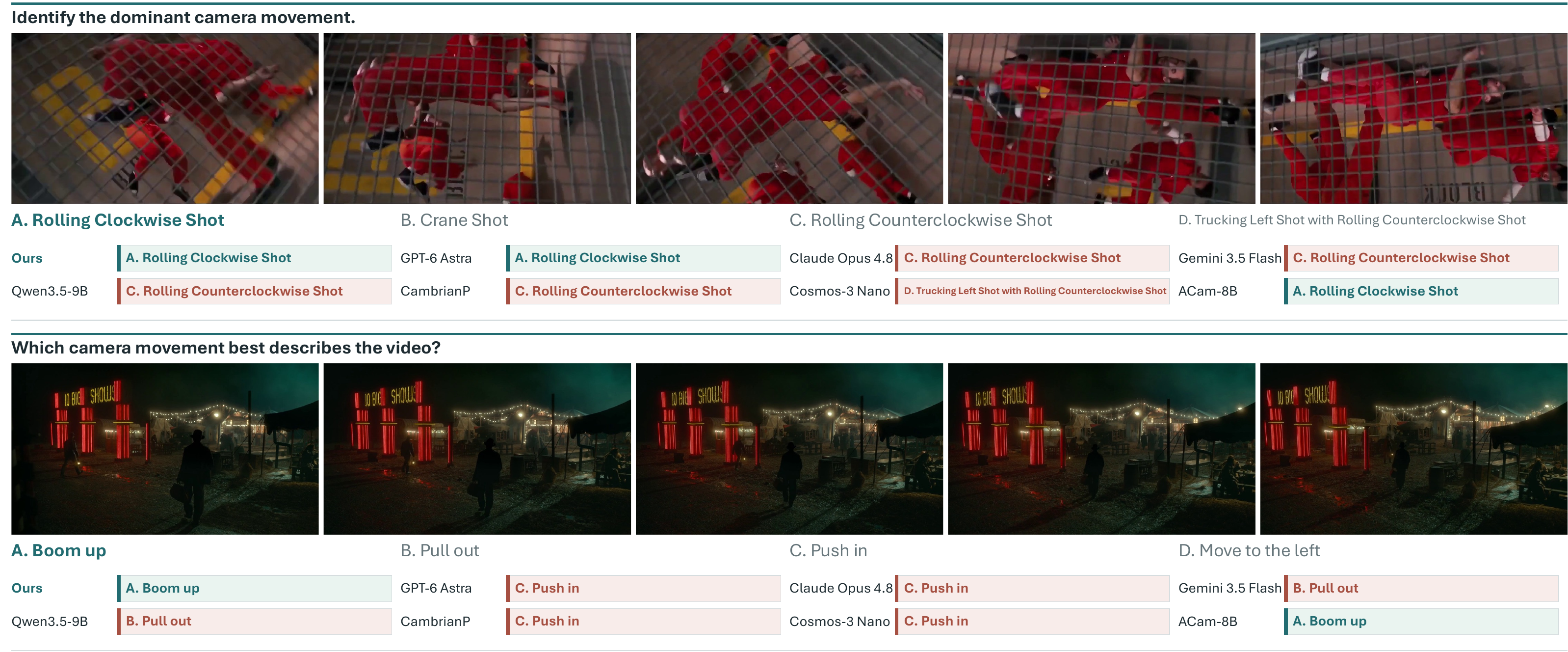}
\caption{Qualitative examples on rolling and boom-up motions. Rolling-clockwise requires distinguishing rotation direction; boom-up is commonly misread as push-in or pull-out by baselines that lack vertical-translation awareness.}
\label{fig:suppqa4}
\end{figure}

Figures~\ref{fig:suppqa1}--\ref{fig:suppqa4} present eight additional
benchmark clips in the same format as Fig.~\ref{fig:qual}: each row shows a
video's sampled frames, the multiple-choice question and ground-truth answer,
followed by every model's prediction colour-coded by correctness. The examples
are drawn from all three main benchmarks and span a range of camera-motion
primitives including pull-out, compound pan-push, tilt, roll, boom-up and
dolly-zoom. Across these cases, most baselines struggle with motions that
require distinguishing translation from zoom or resolving rotation direction,
whereas \method{} consistently produces the correct answer. The full video for
the first clip of Fig.~\ref{fig:qual} (the truck-left shot) is included in the
supplementary material.

\end{document}

%% file: math_commands.tex
\usepackage{amsmath,amsfonts,bm}

\def\eqref#1{equation~\ref{#1}}

\def\1{\bm{1}}

\DeclareMathAlphabet{\mathsfit}{\encodingdefault}{\sfdefault}{m}{sl}
\SetMathAlphabet{\mathsfit}{bold}{\encodingdefault}{\sfdefault}{bx}{n}

%% file: main.bbl
\begin{thebibliography}{77}
\providecommand{\natexlab}[1]{#1}
\providecommand{\url}[1]{\texttt{#1}}
\expandafter\ifx\csname urlstyle\endcsname\relax
  \providecommand{\doi}[1]{doi: #1}\else
  \providecommand{\doi}{doi: \begingroup \urlstyle{rm}\Url}\fi

\bibitem[Bai et~al.(2025{\natexlab{a}})Bai, Cai, Chen, Chen, Chen, Cheng, Deng,
  Ding, Gao, Ge, Ge, Guo, Huang, Huang, Huang, Hui, Jiang, Li, Li, Li, Li, Lin,
  Lin, Liu, Liu, Liu, Liu, Liu, Liu, Lu, Luo, Lv, Men, Meng, Ren, Ren, Song,
  Sun, Tang, Tu, Wan, Wang, Wang, Wang, Wang, Xie, Xu, Xu, Xu, Yang, Yang,
  Yang, Yang, Yu, Zhang, Zhang, Zhang, Zheng, Zhong, Zhou, Zhou, Zhou, Zhu, and
  Zhu]{qwen3vl_tr}
Shuai Bai, Yuxuan Cai, Ruizhe Chen, Keqin Chen, Xionghui Chen, Zesen Cheng,
  Lianghao Deng, Wei Ding, Chang Gao, Chunjiang Ge, Wenbin Ge, Zhifang Guo,
  Qidong Huang, Jie Huang, Fei Huang, Binyuan Hui, Shutong Jiang, Zhaohai Li,
  Mingsheng Li, Mei Li, Kaixin Li, Zicheng Lin, Junyang Lin, Xuejing Liu,
  Jiawei Liu, Chenglong Liu, Yang Liu, Dayiheng Liu, Shixuan Liu, Dunjie Lu,
  Ruilin Luo, Chenxu Lv, Rui Men, Lingchen Meng, Xuancheng Ren, Xingzhang Ren,
  Sibo Song, Yuchong Sun, Jun Tang, Jianhong Tu, Jianqiang Wan, Peng Wang,
  Pengfei Wang, Qiuyue Wang, Yuxuan Wang, Tianbao Xie, Yiheng Xu, Haiyang Xu,
  Jin Xu, Zhibo Yang, Mingkun Yang, Jianxin Yang, An~Yang, Bowen Yu, Fei Zhang,
  Hang Zhang, Xi~Zhang, Bo~Zheng, Humen Zhong, Jingren Zhou, Fan Zhou, Jing
  Zhou, Yuanzhi Zhu, and Ke~Zhu.
\newblock {Qwen3-VL Technical Report}, 2025{\natexlab{a}}.
\newblock URL \url{https://arxiv.org/abs/2511.21631}.

\bibitem[Bai et~al.(2025{\natexlab{b}})Bai, Chen, Liu, Wang, Ge, Song, Dang,
  Wang, Wang, Tang, Zhong, Zhu, Yang, Li, Wan, Wang, Ding, Fu, Xu, Ye, Zhang,
  Xie, Cheng, Zhang, Yang, Xu, and Lin]{qwen25vl}
Shuai Bai, Keqin Chen, Xuejing Liu, Jialin Wang, Wenbin Ge, Sibo Song, Kai
  Dang, Peng Wang, Shijie Wang, Jun Tang, Humen Zhong, Yuanzhi Zhu, Mingkun
  Yang, Zhaohai Li, Jianqiang Wan, Pengfei Wang, Wei Ding, Zheren Fu, Yiheng
  Xu, Jiabo Ye, Xi~Zhang, Tianbao Xie, Zesen Cheng, Hang Zhang, Zhibo Yang,
  Haiyang Xu, and Junyang Lin.
\newblock {Qwen2.5-VL Technical Report}, 2025{\natexlab{b}}.
\newblock URL \url{https://arxiv.org/abs/2502.13923}.

\bibitem[Butler et~al.(2012)Butler, Wulff, Stanley, and Black]{sintel}
Daniel~J. Butler, Jonas Wulff, Garrett~B. Stanley, and Michael~J. Black.
\newblock {A Naturalistic Open Source Movie for Optical Flow Evaluation}.
\newblock In \emph{Computer Vision -- ECCV 2012}, pp.\  611–625. Springer
  Berlin Heidelberg, 2012.
\newblock ISBN 9783642337833.
\newblock \doi{10.1007/978-3-642-33783-3_44}.
\newblock URL \url{http://dx.doi.org/10.1007/978-3-642-33783-3_44}.

\bibitem[Cai et~al.(2025)Cai, Yeh, Xu, Liu, Meyer, Lei, Zhao, Li, Chandra, and
  Shi]{depthlm}
Zhipeng Cai, Ching-Feng Yeh, Hu~Xu, Zhuang Liu, Gregory Meyer, Xinjie Lei,
  Changsheng Zhao, Shang-Wen Li, Vikas Chandra, and Yangyang Shi.
\newblock {DepthLM: Metric Depth From Vision Language Models}, 2025.
\newblock URL \url{https://arxiv.org/abs/2509.25413}.

\bibitem[Chaumette \& Hutchinson(2006)Chaumette and
  Hutchinson]{chaumette2006visual}
Fran{\c{c}}ois Chaumette and Seth Hutchinson.
\newblock {Visual servo control. I. Basic approaches}.
\newblock \emph{IEEE Robotics \& Automation Magazine}, 13\penalty0
  (4):\penalty0 82--90, 2006.
\newblock \doi{10.1109/MRA.2006.250573}.

\bibitem[Chen et~al.(2025)Chen, Zhang, Yu, Luo, Sun, Pan, An, Feng, Pei, Cai,
  and Huang]{thinker3d}
Zhangquan Chen, Manyuan Zhang, Xinlei Yu, Xufang Luo, Mingze Sun, Zihao Pan,
  Xiang An, Yan Feng, Peng Pei, Xunliang Cai, and Ruqi Huang.
\newblock {Think with 3D: Geometric Imagination Grounded Spatial Reasoning from
  Limited Views}, 2025.
\newblock URL \url{https://arxiv.org/abs/2510.18632}.

\bibitem[Cheng et~al.(2025{\natexlab{a}})Cheng, Hu, Yu, Zhao, Wang, and
  Wang]{cheng2025reggsunposedsparseviews}
Chong Cheng, Yu~Hu, Sicheng Yu, Beizhen Zhao, Zijian Wang, and Hao Wang.
\newblock {RegGS: Unposed Sparse Views Gaussian Splatting with 3DGS
  Registration}, 2025{\natexlab{a}}.
\newblock URL \url{https://arxiv.org/abs/2507.08136}.

\bibitem[Cheng et~al.(2025{\natexlab{b}})Cheng, Song, Yao, Zhou, Zhang, and
  Wang]{cheng2025graphguidedscenereconstructionimages}
Chong Cheng, Gaochao Song, Yiyang Yao, Qinzheng Zhou, Gangjian Zhang, and Hao
  Wang.
\newblock {Graph-Guided Scene Reconstruction from Images with 3D Gaussian
  Splatting}, 2025{\natexlab{b}}.
\newblock URL \url{https://arxiv.org/abs/2502.17377}.

\bibitem[Cheng et~al.(2025{\natexlab{c}})Cheng, Wang, Yu, Hu, Yao, and
  Wang]{cheng2025unposed3dgsreconstructionprobabilistic}
Chong Cheng, Zijian Wang, Sicheng Yu, Yu~Hu, Nanjie Yao, and Hao Wang.
\newblock {Unposed 3DGS Reconstruction with Probabilistic Procrustes Mapping},
  2025{\natexlab{c}}.
\newblock URL \url{https://arxiv.org/abs/2507.18541}.

\bibitem[Cheng et~al.(2025{\natexlab{d}})Cheng, Yu, Wang, Zhou, and
  Wang]{cheng2025outdoormonocularslamglobal}
Chong Cheng, Sicheng Yu, Zijian Wang, Yifan Zhou, and Hao Wang.
\newblock {Outdoor Monocular SLAM with Global Scale-Consistent 3D Gaussian
  Pointmaps}, 2025{\natexlab{d}}.
\newblock URL \url{https://arxiv.org/abs/2507.03737}.

\bibitem[Cheng et~al.(2025{\natexlab{e}})Cheng, Zhou, Zhang, Zeng, and
  Wang]{10972824}
Chong Cheng, Qinzheng Zhou, Jianfeng Zhang, Shiya Zeng, and Hao Wang.
\newblock {Exploring Decoupled Generation for Enhancing VR Interaction and
  Control}.
\newblock In \emph{2025 IEEE Conference on Virtual Reality and 3D User
  Interfaces Abstracts and Workshops (VRW)}, pp.\  1328--1329,
  2025{\natexlab{e}}.
\newblock \doi{10.1109/VRW66409.2025.00310}.

\bibitem[Cheng et~al.(2026{\natexlab{a}})Cheng, Chen, Xie, Yin, Ren, Zhang,
  Guo, and Wang]{cheng2026longstreamlongsequencestreamingautoregressive}
Chong Cheng, Xianda Chen, Tao Xie, Wei Yin, Weiqiang Ren, Qian Zhang, Xiaoyang
  Guo, and Hao Wang.
\newblock {LongStream: Long-Sequence Streaming Autoregressive Visual Geometry},
  2026{\natexlab{a}}.
\newblock URL \url{https://arxiv.org/abs/2602.13172}.

\bibitem[Cheng et~al.(2026{\natexlab{b}})Cheng, Tao, Yao, Ding, Chen, Du, Guo,
  Yin, Ren, Zhang, Chen, and
  Wang]{cheng2026horizonstreamlonghorizonattentionstreaming}
Chong Cheng, Peilin Tao, Nanjie Yao, Guanzhi Ding, Xianda Chen, Yuansen Du,
  Xiaoyang Guo, Wei Yin, Weiqiang Ren, Qian Zhang, Zhengqing Chen, and Hao
  Wang.
\newblock {HorizonStream: Long-Horizon Attention for Streaming 3D
  Reconstruction}, 2026{\natexlab{b}}.
\newblock URL \url{https://arxiv.org/abs/2605.23889}.

\bibitem[Cho et~al.(2024)Cho, Huang, Nam, An, Kim, and Lee]{locotrack}
Seokju Cho, Jiahui Huang, Jisu Nam, Honggyu An, Seungryong Kim, and Joon-Young
  Lee.
\newblock {Local All-Pair Correspondence for Point Tracking}, 2024.
\newblock URL \url{https://arxiv.org/abs/2407.15420}.

\bibitem[Courant et~al.(2024)Courant, Dufour, Wang, Christie, and
  Kalogeiton]{courant2024exceptional}
Robin Courant, Nicolas Dufour, Xi~Wang, Marc Christie, and Vicky Kalogeiton.
\newblock Et the exceptional trajectories: Text-to-camera-trajectory generation
  with character awareness.
\newblock In \emph{European conference on computer vision}, pp.\  464--480.
  Springer, 2024.

\bibitem[Dai et~al.(2017)Dai, Chang, Savva, Halber, Funkhouser, and
  Nießner]{scannet}
Angela Dai, Angel~X. Chang, Manolis Savva, Maciej Halber, Thomas Funkhouser,
  and Matthias Nießner.
\newblock {ScanNet: Richly-annotated 3D Reconstructions of Indoor Scenes},
  2017.
\newblock URL \url{https://arxiv.org/abs/1702.04405}.

\bibitem[Dang et~al.(2026)Dang, Guo, Hou, Leng, Li, Li, Liu, Mao, Wang, Yuan,
  Zhu, Lin, Bai, Jiang, Zhao, Zeng, Gao, Jiang, Cen, Huang, Wang, Zhang, Liu,
  Yang, Lu, and Zhao]{rynnbrain_v1}
Ronghao Dang, Jiayan Guo, Bohan Hou, Sicong Leng, Kehan Li, Xin Li, Jiangpin
  Liu, Yunxuan Mao, Zhikai Wang, Yuqian Yuan, Minghao Zhu, Xiao Lin, Yang Bai,
  Qian Jiang, Yaxi Zhao, Minghua Zeng, Junlong Gao, Yuming Jiang, Jun Cen,
  Siteng Huang, Liuyi Wang, Wenqiao Zhang, Chengju Liu, Jianfei Yang, Shijian
  Lu, and Deli Zhao.
\newblock {RynnBrain: Open Embodied Foundation Models}, 2026.
\newblock URL \url{https://arxiv.org/abs/2602.14979}.

\bibitem[Doersch et~al.(2022)Doersch, Gupta, Markeeva, Recasens, Smaira, Aytar,
  Carreira, Zisserman, and Yang]{tapvid}
Carl Doersch, Ankush Gupta, Larisa Markeeva, Adrià Recasens, Lucas Smaira,
  Yusuf Aytar, João Carreira, Andrew Zisserman, and Yi~Yang.
\newblock {TAP-Vid: A Benchmark for Tracking Any Point in a Video}, 2022.
\newblock URL \url{https://arxiv.org/abs/2211.03726}.

\bibitem[Doersch et~al.(2023)Doersch, Yang, Vecerik, Gokay, Gupta, Aytar,
  Carreira, and Zisserman]{tapir}
Carl Doersch, Yi~Yang, Mel Vecerik, Dilara Gokay, Ankush Gupta, Yusuf Aytar,
  Joao Carreira, and Andrew Zisserman.
\newblock {TAPIR: Tracking Any Point with per-frame Initialization and temporal
  Refinement}, 2023.
\newblock URL \url{https://arxiv.org/abs/2306.08637}.

\bibitem[Doersch et~al.(2024)Doersch, Luc, Yang, Gokay, Koppula, Gupta,
  Heyward, Rocco, Goroshin, Carreira, and Zisserman]{bootstapir}
Carl Doersch, Pauline Luc, Yi~Yang, Dilara Gokay, Skanda Koppula, Ankush Gupta,
  Joseph Heyward, Ignacio Rocco, Ross Goroshin, Jo{\~a}o Carreira, and Andrew
  Zisserman.
\newblock {BootsTAP: Bootstrapped Training for Tracking-Any-Point}, 2024.
\newblock URL \url{https://arxiv.org/abs/2402.00847}.

\bibitem[Du et~al.(2026)Du, Du, Liu, Yu, Gu, Han, Liu, Liu, Zhang, Chen, and
  Guo]{camdistill}
Dazhao Du, Shiyan Du, Jian Liu, Yongjian Yu, Bohai Gu, Tao Han, Hualuo Liu,
  Eric Liu, Yujia Zhang, Xi~Chen, and Song Guo.
\newblock {Temporally Grounded Compositional Camera Motion Understanding via
  Geometric Knowledge Distillation}, 2026.
\newblock URL \url{https://arxiv.org/abs/2608.10932}.

\bibitem[Fan et~al.(2025)Fan, Zhang, Li, Zhang, Chen, Hu, Wang, Qu, Zhou, Wang,
  Yan, Xu, Theiss, Chen, Li, Tu, Wang, and Ranjan]{vlm3r}
Zhiwen Fan, Jian Zhang, Renjie Li, Junge Zhang, Runjin Chen, Hezhen Hu, Kevin
  Wang, Huaizhi Qu, Shijie Zhou, Dilin Wang, Zhicheng Yan, Hongyu Xu, Justin
  Theiss, Tianlong Chen, Jiachen Li, Zhengzhong Tu, Zhangyang Wang, and Rakesh
  Ranjan.
\newblock {VLM-3R: Vision-Language Models Augmented with Instruction-Aligned 3D
  Reconstruction}, 2025.
\newblock URL \url{https://arxiv.org/abs/2505.20279}.

\bibitem[Feng et~al.(2026)Feng, Musunuri, and Su]{geoguided}
Haoan Feng, Sri~Harsha Musunuri, and Guan-Ming Su.
\newblock {Geometry-Guided Camera Motion Understanding in VideoLLMs}, 2026.
\newblock URL \url{https://arxiv.org/abs/2603.13119}.

\bibitem[Fu et~al.(2024)Fu, Dai, Luo, Li, Ren, Zhang, Wang, Zhou, Shen, Zhang,
  Chen, Li, Lin, Zhao, Li, Xu, Zheng, Chen, Shan, He, and Sun]{videomme}
Chaoyou Fu, Yuhan Dai, Yongdong Luo, Lei Li, Shuhuai Ren, Renrui Zhang, Zihan
  Wang, Chenyu Zhou, Yunhang Shen, Mengdan Zhang, Peixian Chen, Yanwei Li,
  Shaohui Lin, Sirui Zhao, Ke~Li, Tong Xu, Xiawu Zheng, Enhong Chen, Caifeng
  Shan, Ran He, and Xing Sun.
\newblock {Video-MME: The First-Ever Comprehensive Evaluation Benchmark of
  Multi-modal LLMs in Video Analysis}, 2024.
\newblock URL \url{https://arxiv.org/abs/2405.21075}.

\bibitem[Gao et~al.(2026)Gao, Chen, Qu, Song, Wang, Zhao, Wang, and
  Li]{qgeomem}
Xianqiang Gao, Qizhi Chen, Delin Qu, Haoming Song, Zhigang Wang, Bin Zhao, Dong
  Wang, and Xuelong Li.
\newblock {Q-GeoMem: Question-Guided Geometric Memory for Video Spatial
  Reasoning}, 2026.
\newblock URL \url{https://arxiv.org/abs/2605.27318}.

\bibitem[Greff et~al.(2022)Greff, Belletti, Beyer, Doersch, Du, Duckworth,
  Fleet, Gnanapragasam, Golemo, Herrmann, Kipf, Kundu, Lagun, Laradji, Liu,
  Meyer, Miao, Nowrouzezahrai, Oztireli, Pot, Radwan, Rebain, Sabour, Sajjadi,
  Sela, Sitzmann, Stone, Sun, Vora, Wang, Wu, Yi, Zhong, and
  Tagliasacchi]{kubric}
Klaus Greff, Francois Belletti, Lucas Beyer, Carl Doersch, Yilun Du, Daniel
  Duckworth, David~J. Fleet, Dan Gnanapragasam, Florian Golemo, Charles
  Herrmann, Thomas Kipf, Abhijit Kundu, Dmitry Lagun, Issam Laradji,
  Hsueh-Ti~Derek Liu, Henning Meyer, Yishu Miao, Derek Nowrouzezahrai, Cengiz
  Oztireli, Etienne Pot, Noha Radwan, Daniel Rebain, Sara Sabour, Mehdi S.~M.
  Sajjadi, Matan Sela, Vincent Sitzmann, Austin Stone, Deqing Sun, Suhani Vora,
  Ziyu Wang, Tianhao Wu, Kwang~Moo Yi, Fangcheng Zhong, and Andrea
  Tagliasacchi.
\newblock {Kubric: A scalable dataset generator}, 2022.
\newblock URL \url{https://arxiv.org/abs/2203.03570}.

\bibitem[Harley et~al.(2025)Harley, You, Sun, Zheng, Raghuraman, Gu, Liang,
  Chu, Dave, Tokmakov, You, Ambrus, Fragkiadaki, and Guibas]{alltracker}
Adam~W. Harley, Yang You, Xinglong Sun, Yang Zheng, Nikhil Raghuraman, Yunqi
  Gu, Sheldon Liang, Wen-Hsuan Chu, Achal Dave, Pavel Tokmakov, Suya You, Rares
  Ambrus, Katerina Fragkiadaki, and Leonidas~J. Guibas.
\newblock {AllTracker: Efficient Dense Point Tracking at High Resolution},
  2025.
\newblock URL \url{https://arxiv.org/abs/2506.07310}.

\bibitem[Hu et~al.(2025{\natexlab{a}})Hu, Lin, Long, Ran, Jiang, Wang, Zhu, Xu,
  Wang, and Pang]{g2vlm}
Wenbo Hu, Jingli Lin, Yilin Long, Yunlong Ran, Lihan Jiang, Yifan Wang,
  Chenming Zhu, Runsen Xu, Tai Wang, and Jiangmiao Pang.
\newblock {G$^2$VLM: Geometry Grounded Vision Language Model with Unified 3D
  Reconstruction and Spatial Reasoning}, 2025{\natexlab{a}}.
\newblock URL \url{https://arxiv.org/abs/2511.21688}.

\bibitem[Hu et~al.(2025{\natexlab{b}})Hu, Cheng, Yu, Guo, and
  Wang]{hu2025vggt4dminingmotioncues}
Yu~Hu, Chong Cheng, Sicheng Yu, Xiaoyang Guo, and Hao Wang.
\newblock {VGGT4D: Mining Motion Cues in Visual Geometry Transformers for 4D
  Scene Reconstruction}, 2025{\natexlab{b}}.
\newblock URL \url{https://arxiv.org/abs/2511.19971}.

\bibitem[Huang et~al.(2025)Huang, Zhou, Rabeti, Korovko, Ling, Ren, Shen, Gao,
  Slepichev, Lin, Ren, Xie, Biswas, Leal-Taixe, and Fidler]{vipe}
Jiahui Huang, Qunjie Zhou, Hesam Rabeti, Aleksandr Korovko, Huan Ling, Xuanchi
  Ren, Tianchang Shen, Jun Gao, Dmitry Slepichev, Chen-Hsuan Lin, Jiawei Ren,
  Kevin Xie, Joydeep Biswas, Laura Leal-Taixe, and Sanja Fidler.
\newblock {ViPE: Video Pose Engine for 3D Geometric Perception}, 2025.
\newblock URL \url{https://arxiv.org/abs/2508.10934}.

\bibitem[Karaev et~al.(2023)Karaev, Rocco, Graham, Neverova, Vedaldi, and
  Rupprecht]{cotracker}
Nikita Karaev, Ignacio Rocco, Benjamin Graham, Natalia Neverova, Andrea
  Vedaldi, and Christian Rupprecht.
\newblock {CoTracker: It is Better to Track Together}, 2023.
\newblock URL \url{https://arxiv.org/abs/2307.07635}.

\bibitem[Karaev et~al.(2024)Karaev, Makarov, Wang, Neverova, Vedaldi, and
  Rupprecht]{cotracker3}
Nikita Karaev, Iurii Makarov, Jianyuan Wang, Natalia Neverova, Andrea Vedaldi,
  and Christian Rupprecht.
\newblock {CoTracker3: Simpler and Better Point Tracking by Pseudo-Labelling
  Real Videos}, 2024.
\newblock URL \url{https://arxiv.org/abs/2410.11831}.

\bibitem[Keetha et~al.(2025)Keetha, M{\"u}ller, Sch{\"o}nberger, Porzi, Zhang,
  Fischer, Knapitsch, Zauss, Weber, Antunes, Luiten, Lopez-Antequera, Bul{\`o},
  Richardt, Ramanan, Scherer, and Kontschieder]{mapanything}
Nikhil Keetha, Norman M{\"u}ller, Johannes Sch{\"o}nberger, Lorenzo Porzi,
  Yuchen Zhang, Tobias Fischer, Arno Knapitsch, Duncan Zauss, Ethan Weber,
  Nelson Antunes, Jonathon Luiten, Manuel Lopez-Antequera, Samuel~Rota
  Bul{\`o}, Christian Richardt, Deva Ramanan, Sebastian Scherer, and Peter
  Kontschieder.
\newblock {MapAnything: Universal Feed-Forward Metric 3D Reconstruction}, 2025.
\newblock URL \url{https://arxiv.org/abs/2509.13414}.

\bibitem[Kong et~al.(2025)Kong, Duan, Xu, Guo, Zhu, and Shi]{lrrbench}
Fei Kong, Jinhao Duan, Kaidi Xu, Zhenhua Guo, Xiaofeng Zhu, and Xiaoshuang Shi.
\newblock {LRR-Bench: Left, Right or Rotate? Vision-Language models Still
  Struggle With Spatial Understanding Tasks}, 2025.
\newblock URL \url{https://arxiv.org/abs/2507.20174}.

\bibitem[Lai \& Vedaldi(2025)Lai and Vedaldi]{tracktention}
Zihang Lai and Andrea Vedaldi.
\newblock {Tracktention: Leveraging Point Tracking to Attend Videos Faster and
  Better}, 2025.
\newblock URL \url{https://arxiv.org/abs/2503.19904}.

\bibitem[Lai et~al.(2026)Lai, Insafutdinov, Sucar, and Vedaldi]{cowtracker}
Zihang Lai, Eldar Insafutdinov, Edgar Sucar, and Andrea Vedaldi.
\newblock {CoWTracker: Tracking by Warping instead of Correlation}, 2026.
\newblock URL \url{https://arxiv.org/abs/2602.04877}.

\bibitem[Leroy et~al.(2024)Leroy, Cabon, and Revaud]{mast3r}
Vincent Leroy, Yohann Cabon, and J{\'e}r{\^o}me Revaud.
\newblock {Grounding Image Matching in 3D with MASt3R}, 2024.
\newblock URL \url{https://arxiv.org/abs/2406.09756}.

\bibitem[Li et~al.(2024)Li, Zhang, Liu, Zeng, Ren, Li, and Zhang]{taptr}
Hongyang Li, Hao Zhang, Shilong Liu, Zhaoyang Zeng, Tianhe Ren, Feng Li, and
  Lei Zhang.
\newblock {TAPTR: Tracking Any Point with Transformers as Detection}, 2024.
\newblock URL \url{https://arxiv.org/abs/2403.13042}.

\bibitem[Li et~al.(2026)Li, Hou, Zhu, Zhang, Cheng, Wang, Leng, Li, Lin, Yao,
  Zeng, Liu, Dang, Guo, Huang, Zhao, Ping, Zhao, Zhao, Wang, Lu, Xue, Tang,
  Wang, Wang, Gao, Lu, Liu, Yang, Chen, and Zhao]{rynnbrain}
Kehan Li, Bohan Hou, Minghao Zhu, Tianyi Zhang, Zesen Cheng, Zhikai Wang,
  Sicong Leng, Xin Li, Xiao Lin, Biying Yao, Minghua Zeng, Jiangpin Liu,
  Ronghao Dang, Jiayan Guo, Siteng Huang, Haoyu Zhao, Heng Ping, Yaxi Zhao,
  Tong Zhao, Kexiang Wang, Tong Lu, Shengke Xue, Jiahao Tang, Yulei Wang,
  Zejing Wang, Jianwei Gao, Shijian Lu, Chengju Liu, Jianfei Yang, Mingxiu
  Chen, and Deli Zhao.
\newblock {RynnBrain 1.1: Towards More Capable and Generalizable Embodied
  Foundation Model}, 2026.
\newblock URL \url{https://arxiv.org/abs/2607.17977}.

\bibitem[Li et~al.(2025)Li, Wang, Yu, Zeng, Zhu, Huang, Gao, Li, He, Wang,
  Qiao, Wang, and Wang]{videochatflash}
Xinhao Li, Yi~Wang, Jiashuo Yu, Xiangyu Zeng, Yuhan Zhu, Haian Huang, Jianfei
  Gao, Kunchang Li, Yinan He, Chenting Wang, Yu~Qiao, Yali Wang, and Limin
  Wang.
\newblock {VideoChat-Flash: Hierarchical Compression for Long-Context Video
  Modeling}, 2025.
\newblock URL \url{https://arxiv.org/abs/2501.00574}.

\bibitem[Lin et~al.(2025)Lin, Cen, Jiang, Karhade, Wang, Mitra, Ling, Huang,
  Liu, Chen, Zawar, Bai, Du, Gan, and Ramanan]{camerabench}
Zhiqiu Lin, Siyuan Cen, Daniel Jiang, Jay Karhade, Hewei Wang, Chancharik
  Mitra, Tiffany Ling, Yuhan Huang, Sifan Liu, Mingyu Chen, Rushikesh Zawar,
  Xue Bai, Yilun Du, Chuang Gan, and Deva Ramanan.
\newblock {Towards Understanding Camera Motions in Any Video}, 2025.
\newblock URL \url{https://arxiv.org/abs/2504.15376}.

\bibitem[Liu et~al.(2025)Liu, He, Jin, Zheng, Dong, Zhang, Huang, He, Li, Chen,
  Qiao, Ouyang, Zhao, and Liu]{shotbench}
Hongbo Liu, Jingwen He, Yi~Jin, Dian Zheng, Yuhao Dong, Fan Zhang, Ziqi Huang,
  Yinan He, Yangguang Li, Weichao Chen, Yu~Qiao, Wanli Ouyang, Shengjie Zhao,
  and Ziwei Liu.
\newblock {ShotBench: Expert-Level Cinematic Understanding in Vision-Language
  Models}, 2025.
\newblock URL \url{https://arxiv.org/abs/2506.21356}.

\bibitem[Lu et~al.(2023)Lu, Bansal, Xia, Liu, Li, Hajishirzi, Cheng, Chang,
  Galley, and Gao]{mathvista}
Pan Lu, Hritik Bansal, Tony Xia, Jiacheng Liu, Chunyuan Li, Hannaneh
  Hajishirzi, Hao Cheng, Kai-Wei Chang, Michel Galley, and Jianfeng Gao.
\newblock {MathVista: Evaluating Mathematical Reasoning of Foundation Models in
  Visual Contexts}, 2023.
\newblock URL \url{https://arxiv.org/abs/2310.02255}.

\bibitem[Moing et~al.(2023)Moing, Ponce, and Schmid]{dot_tracking}
Guillaume~Le Moing, Jean Ponce, and Cordelia Schmid.
\newblock {Dense Optical Tracking: Connecting the Dots}, 2023.
\newblock URL \url{https://arxiv.org/abs/2312.00786}.

\bibitem[Ngo et~al.(2024)Ngo, Zhuang, Gan, Kalogerakis, Tulyakov, Lee, and
  Wang]{delta_track}
Tuan~Duc Ngo, Peiye Zhuang, Chuang Gan, Evangelos Kalogerakis, Sergey Tulyakov,
  Hsin-Ying Lee, and Chaoyang Wang.
\newblock {DELTA: Dense Efficient Long-range 3D Tracking for any video}, 2024.
\newblock URL \url{https://arxiv.org/abs/2410.24211}.

\bibitem[{NVIDIA} et~al.(2025){NVIDIA}, Agarwal, Ali, Bala, Balaji, Barker,
  Cai, Chattopadhyay, Chen, Cui, Ding, Dworakowski, Fan, Fenzi, Ferroni,
  Fidler, Fox, Ge, Ge, Gu, Gururani, He, Huang, Huffman, Jannaty, Jin, Kim,
  Klár, Lam, Lan, Leal-Taixe, Li, Li, Lin, Lin, Ling, Liu, Liu, Luo, Ma, Mao,
  Mo, Mousavian, Nah, Niverty, Page, Paschalidou, Patel, Pavao, Ramezanali,
  Reda, Ren, Sabavat, Schmerling, Shi, Stefaniak, Tang, Tchapmi, Tredak, Tseng,
  Varghese, Wang, Wang, Wang, Wang, Wei, Wei, Wu, Xu, Yang, Yen-Chen, Zeng,
  Zeng, Zhang, Zhang, Zhang, Zhao, and Zolkowski]{cosmos1}
{NVIDIA}, Niket Agarwal, Arslan Ali, Maciej Bala, Yogesh Balaji, Erik Barker,
  Tiffany Cai, Prithvijit Chattopadhyay, Yongxin Chen, Yin Cui, Yifan Ding,
  Daniel Dworakowski, Jiaojiao Fan, Michele Fenzi, Francesco Ferroni, Sanja
  Fidler, Dieter Fox, Songwei Ge, Yunhao Ge, Jinwei Gu, Siddharth Gururani,
  Ethan He, Jiahui Huang, Jacob Huffman, Pooya Jannaty, Jingyi Jin, Seung~Wook
  Kim, Gergely Klár, Grace Lam, Shiyi Lan, Laura Leal-Taixe, Anqi Li, Zhaoshuo
  Li, Chen-Hsuan Lin, Tsung-Yi Lin, Huan Ling, Ming-Yu Liu, Xian Liu, Alice
  Luo, Qianli Ma, Hanzi Mao, Kaichun Mo, Arsalan Mousavian, Seungjun Nah,
  Sriharsha Niverty, David Page, Despoina Paschalidou, Zeeshan Patel, Lindsey
  Pavao, Morteza Ramezanali, Fitsum Reda, Xiaowei Ren, Vasanth Rao~Naik
  Sabavat, Ed~Schmerling, Stella Shi, Bartosz Stefaniak, Shitao Tang, Lyne
  Tchapmi, Przemek Tredak, Wei-Cheng Tseng, Jibin Varghese, Hao Wang, Haoxiang
  Wang, Heng Wang, Ting-Chun Wang, Fangyin Wei, Xinyue Wei, Jay~Zhangjie Wu,
  Jiashu Xu, Wei Yang, Lin Yen-Chen, Xiaohui Zeng, Yu~Zeng, Jing Zhang,
  Qinsheng Zhang, Yuxuan Zhang, Qingqing Zhao, and Artur Zolkowski.
\newblock {Cosmos World Foundation Model Platform for Physical AI}, 2025.
\newblock URL \url{https://arxiv.org/abs/2501.03575}.

\bibitem[{NVIDIA} et~al.(2026){NVIDIA}, {Aditi}, Agarwal, Ali, Allen, Antolini,
  Aubame, Azzolini, Bai, Bala, Balaji, Bapst, Basant, Beladiya, Bhat, Bhat,
  Blick, Brighella, Cai, Cai, Cameracci, Cao, Cao, Carlson, Casanova, Chang,
  Chang, Chao, Chattopadhyay, Chaudhari, Chen, Chen, Chen, Chen, Chen, Chen,
  Chen, Cheng, Cheng, Chia, Choi, Chung, Cong, Cui, Dadela, Dadhich, Dai, Daw,
  Degirmenci, Del~Monte, Denomme, Dharur, Di~Lucca, Ding, Ding, Ding, Dong,
  Drumheller, Du, Dzhumamuratova, Efitorov, Eghbalzadeh, Eigbe, Hanafi, Eslami,
  Falk, Fan, Fan, Fasale, Fefilatyev, Feng, Ferroni, Fidler, Fu, Fugro,
  Gaikwad, Galda, Gao, Gao, Ge, Ghosh, Goel, Goel, Gokul, Govindaraju, Gu,
  Guerrero, Guo, Gupta, Gururani, Hadfield, Han, Handa, Hao, Harrim, Hassani,
  Hayes-Roth, He, Helvig, Hogg, Huang, Huang, Huang, Huang, Huffman, Hutchins,
  Indupuru, Ivanovic, Jain, Jang, Ji, Jian, Jiang, Jin, Joshi, Joshi, Joshi,
  Ju, Jung, Kang, Kassekert, Kautz, Khetan, Kiczka, Kierat, Kim, Kim, Kim,
  Kong, Kong, Kong, Kornuta, Krivov, Kuang, Kumar, Kuo, Kurian, Kutak,
  Lafleche, Lahkar, Laymoun, Lee, Lee, Leone, Li, Li, Li, Li, Li, Li, Li, Li,
  Li, Li, Li, Li, Liang, Liao, Lin, Lin, Liu, Liu, Liu, Lu, Lu, Luo, Luo, Luo,
  Lyu, Ma, Ma, Ma, Majchrowski, Marcoux, Martin, Miao, Mirzaei, Misra, Mo,
  Mohsin, Moon, Morkisz, Motiian, Motkov, Nah, Narang, Narayanan, Ngazimbi,
  Ouyang, Pachori, Page, Pang, Park, Patekar, Patwary, Pavone, Pham, Ping,
  Pouya, Prabhumoye, Praveen, Qu, Rabeti, Ramezanali, Reeb, Ren, Rumley, Rymer,
  Saito, Seol, Shao, Shekdar, Shen, Shi, Shi, Shi, Shih, Shoeybi, Sieniawski,
  Song, Sotelo, Sotoodeh, Srinivasa, Srinivasakumar, Stefaniak, Steiger, Sun,
  Tang, Tang, Tang, Tang, Tavakkoli, Ting, Tomala, Tseng, Varghese, Vasilev,
  Volk, Wagwani, Waleffe, Wang, Wang, Wang, Wang, Wang, Wang, Wang, Wang, Wang,
  Watve, Wehr, Wei, Weng, Wu, Wu, Xia, Xiao, Xiao, Xie, Xu, Xu, Xu, Xu, Xu, Xu,
  Yang, Yang, Yang, Yang, Yang, Yang, You, Yu, Yuan, Yuen, Zeng, Zeren, Zha,
  Zhang, Zhang, Zhang, Zhang, Zhang, Zhang, Zhang, Zhang, Zhao, Zhao,
  Zhautouskaya, Zhou, Zhou, Zhu, Zhu, Zhylko, and Zolkowski]{cosmos3}
{NVIDIA}, {Aditi}, Niket Agarwal, Arslan Ali, Jon Allen, Martin Antolini,
  Adeline Aubame, Alisson Azzolini, Junjie Bai, Maciej Bala, Yogesh Balaji,
  Josh Bapst, Aarti Basant, Mukesh Beladiya, Mohammad~Qazim Bhat, Zaid~Pervaiz
  Bhat, Dan Blick, Vanni Brighella, Han Cai, Tiffany Cai, Eric Cameracci,
  Jiaxin Cao, Yulong Cao, Mark Carlson, Carlos Casanova, Ting-Yun Chang, Yan
  Chang, Yu-Wei Chao, Prithvijit Chattopadhyay, Roshan Chaudhari, Chieh-Yun
  Chen, Junyu Chen, Ke~Chen, Qizhi Chen, Wenkai Chen, Xiaotong Chen, Yu~Chen,
  An-Chieh Cheng, Click Cheng, Xiu Chia, Jeana Choi, Chaeyeon Chung, Wenyan
  Cong, Yin Cui, Magdalena Dadela, Nalin Dadhich, Wenliang Dai, Joyjit Daw,
  Alperen Degirmenci, Rodrigo~Vieira Del~Monte, Robert Denomme, Sameer Dharur,
  Marco Di~Lucca, Ke~Ding, Wenhao Ding, Yifan Ding, Yuzhu Dong, Nicole
  Drumheller, Yilun Du, Aigul Dzhumamuratova, Aleksandr Efitorov, Hamid
  Eghbalzadeh, Naomi Eigbe, Imad~El Hanafi, Hassan Eslami, Benedikt Falk,
  Jiaojiao Fan, Jim Fan, Amol Fasale, Sergiy Fefilatyev, Liang Feng, Francesco
  Ferroni, Sanja Fidler, Xiao Fu, Vikram Fugro, Prashant Gaikwad, TJ~Galda,
  Katelyn Gao, Yihuai Gao, Wenhang Ge, Sreyan Ghosh, Arushi Goel, Vivek Goel,
  Akash Gokul, Rama Govindaraju, Jinwei Gu, Miguel Guerrero, Elfie Guo, Aryaman
  Gupta, Siddharth Gururani, Hugo Hadfield, Song Han, Ankur Handa, Zekun Hao,
  Mohammad Harrim, Ali Hassani, Nathan Hayes-Roth, Yufan He, Chris Helvig,
  Cyrus Hogg, Madison Huang, Michael Huang, Sophia Huang, Yufan Huang, Jacob
  Huffman, DeLesley Hutchins, Suneel Indupuru, Boris Ivanovic, Arihant Jain,
  Joel Jang, Ryan Ji, Yanan Jian, Dongfu Jiang, Jingyi Jin, Atharva Joshi,
  Nikhilesh Joshi, Pranjali Joshi, Andy Ju, Jaehun Jung, Weiwei Kang, Scott
  Kassekert, Jan Kautz, Ashna Khetan, Julia Kiczka, Slawek Kierat, Gwanghyun
  Kim, Kuno Kim, Sunny Kim, Kezhi Kong, Xin Kong, Zhifeng Kong, Tomasz Kornuta,
  Egor Krivov, Hui Kuang, Saurav Kumar, Chia-Wen Kuo, George Kurian, Wojciech
  Kutak, JF~Lafleche, Himangshu Lahkar, Omar Laymoun, Jayjun Lee, Sanggil Lee,
  Gabriele Leone, Boyi Li, Freya Li, Jiajun Li, Jinfeng Li, Ling Li, Pengcheng
  Li, Shangru Li, Tingle Li, Xiaolong Li, Xuan Li, Zhaoshuo Li, Zhiqi Li, Hao
  Liang, Maosheng Liao, Chen-Hsuan Lin, Tsung-Yi Lin, Ming-Yu Liu, Sifei Liu,
  Zihan Liu, Hai~Loc Lu, Xiangyu Lu, Alice Luo, Ruipu Luo, Wenjie Luo, Jiangran
  Lyu, Martin~Ding Ma, Nic Ma, Qianli Ma, Dawid Majchrowski, Louis Marcoux,
  Miguel Martin, Qing Miao, Ashkan Mirzaei, Shreyas Misra, Kaichun Mo, Durra
  Mohsin, Hyejin Moon, Pawel Morkisz, Saeid Motiian, Kirill Motkov, Seungjun
  Nah, Yashraj Narang, Deepak Narayanan, Thabang Ngazimbi, Julian Ouyang,
  Shubham Pachori, David Page, Yatian Pang, Sehwi Park, Mahesh Patekar, Mostofa
  Patwary, Marco Pavone, Trung Pham, Wei Ping, Soha Pouya, Shrimai Prabhumoye,
  Varun Praveen, Delin Qu, Hesam Rabeti, Morteza Ramezanali, Marilyn Reeb,
  Xuanchi Ren, Kristen Rumley, Wojciech Rymer, Jun Saito, Yeongho Seol, John
  Shao, Piyush Shekdar, Tianwei Shen, Humphrey Shi, Min Shi, Stella Shi, Kevin
  Shih, Mohammad Shoeybi, Mateusz Sieniawski, Shuran Song, Alexander Sotelo,
  Amir Sotoodeh, Sunil Srinivasa, Vignesh Srinivasakumar, Bartosz Stefaniak,
  Rahul~Heinrich Steiger, Shangkun Sun, Jiaxiang Tang, Shitao Tang, Yangyang
  Tang, Yue Tang, Tolou Tavakkoli, Kayley Ting, Krzysztof Tomala, Wei-Cheng
  Tseng, Jibin Varghese, Sergei Vasilev, Thomas Volk, Raju Wagwani, Roger
  Waleffe, Andrew~Z. Wang, Boxiang Wang, Haoxiang Wang, Qiao Wang, Shihao Wang,
  Shijie Wang, Ting-Chun Wang, Yan Wang, Yu~Wang, Rohit Watve, David Wehr,
  Fangyin Wei, Xinshuo Weng, Jay~Zhangjie Wu, Kedi Wu, Hongchi Xia, Summer
  Xiao, Tianjun Xiao, Kevin Xie, Daguang Xu, Jiashu Xu, Mengyao Xu, Ruqing Xu,
  Xingqian Xu, Yao Xu, Dinghao Yang, Dong Yang, Hans Yang, Xiaodong Yang,
  Xuning Yang, Yichu Yang, Yurong You, Zhiding Yu, Hao Yuan, Simon Yuen,
  Xiaohui Zeng, Pengcuo Zeren, Cindy Zha, Haotian Zhang, Jenny Zhang, Jing
  Zhang, Liangkai Zhang, Paris Zhang, Shun Zhang, Xuanmeng Zhang, Zhizheng
  Zhang, Ann Zhao, Yilin Zhao, Yuliya Zhautouskaya, Charles Zhou, Fengzhe Zhou,
  Shilin Zhu, Yuke Zhu, Dima Zhylko, and Artur Zolkowski.
\newblock {Cosmos 3: Omnimodal World Models for Physical AI}, 2026.
\newblock URL \url{https://arxiv.org/abs/2606.02800}.

\bibitem[Pont-Tuset et~al.(2017)Pont-Tuset, Perazzi, Caelles, Arbeláez,
  Sorkine-Hornung, and Van~Gool]{davis}
Jordi Pont-Tuset, Federico Perazzi, Sergi Caelles, Pablo Arbeláez, Alex
  Sorkine-Hornung, and Luc Van~Gool.
\newblock {The 2017 DAVIS Challenge on Video Object Segmentation}, 2017.
\newblock URL \url{https://arxiv.org/abs/1704.00675}.

\bibitem[{Qwen Team}(2026)]{qwen35}
{Qwen Team}.
\newblock {Qwen3.5: Towards Native Multimodal Agents}, February 2026.
\newblock URL \url{https://qwen.ai/blog?id=qwen3.5}.

\bibitem[Song et~al.(2024)Song, Cheng, and
  Wang]{song2024gvkfgaussianvoxelkernel}
Gaochao Song, Chong Cheng, and Hao Wang.
\newblock {GVKF: Gaussian Voxel Kernel Functions for Highly Efficient Surface
  Reconstruction in Open Scenes}, 2024.
\newblock URL \url{https://arxiv.org/abs/2411.01853}.

\bibitem[Song et~al.(2025)Song, Ou, Kong, Li, Neubig, and Yue]{visualpuzzles}
Yueqi Song, Tianyue Ou, Yibo Kong, Zecheng Li, Graham Neubig, and Xiang Yue.
\newblock {VisualPuzzles: Decoupling Multimodal Reasoning Evaluation from
  Domain Knowledge}, 2025.
\newblock URL \url{https://arxiv.org/abs/2504.10342}.

\bibitem[Sturm et~al.(2012)Sturm, Engelhard, Endres, Burgard, and
  Cremers]{tumrgbd}
J{\"u}rgen Sturm, Nikolas Engelhard, Felix Endres, Wolfram Burgard, and Daniel
  Cremers.
\newblock {A benchmark for the evaluation of RGB-D SLAM systems}.
\newblock In \emph{2012 IEEE/RSJ International Conference on Intelligent Robots
  and Systems}, pp.\  573–580. IEEE, October 2012.
\newblock \doi{10.1109/iros.2012.6385773}.
\newblock URL \url{http://dx.doi.org/10.1109/IROS.2012.6385773}.

\bibitem[Tan et~al.(2026)Tan, Huang, Huang, Li, and Gong]{acam}
Yuwen Tan, Joey Huang, Jin Huang, Haoxiang Li, and Boqing Gong.
\newblock {Natural Language Camera Movement Understanding}, 2026.
\newblock URL \url{https://arxiv.org/abs/2607.03043}.

\bibitem[Wang \& Agapito(2024)Wang and Agapito]{spann3r}
Hengyi Wang and Lourdes Agapito.
\newblock {3D Reconstruction with Spatial Memory}, 2024.
\newblock URL \url{https://arxiv.org/abs/2408.16061}.

\bibitem[Wang et~al.(2025{\natexlab{a}})Wang, Chen, Karaev, Vedaldi, Rupprecht,
  and Novotny]{vggt}
Jianyuan Wang, Minghao Chen, Nikita Karaev, Andrea Vedaldi, Christian
  Rupprecht, and David Novotny.
\newblock {VGGT: Visual Geometry Grounded Transformer}, 2025{\natexlab{a}}.
\newblock URL \url{https://arxiv.org/abs/2503.11651}.

\bibitem[Wang et~al.(2026{\natexlab{a}})Wang, Chen, Zhang, Karaev,
  Schönberger, Labatut, Bojanowski, Novotny, Vedaldi, and
  Rupprecht]{vggtomega}
Jianyuan Wang, Minghao Chen, Shangzhan Zhang, Nikita Karaev, Johannes
  Schönberger, Patrick Labatut, Piotr Bojanowski, David Novotny, Andrea
  Vedaldi, and Christian Rupprecht.
\newblock {VGGT-$\Omega$}, 2026{\natexlab{a}}.
\newblock URL \url{https://arxiv.org/abs/2605.15195}.

\bibitem[Wang et~al.(2025{\natexlab{b}})Wang, Zhang, Holynski, Efros, and
  Kanazawa]{cut3r}
Qianqian Wang, Yifei Zhang, Aleksander Holynski, Alexei~A. Efros, and Angjoo
  Kanazawa.
\newblock {Continuous 3D Perception Model with Persistent State},
  2025{\natexlab{b}}.
\newblock URL \url{https://arxiv.org/abs/2501.12387}.

\bibitem[Wang et~al.(2023)Wang, Leroy, Cabon, Chidlovskii, and Revaud]{dust3r}
Shuzhe Wang, Vincent Leroy, Yohann Cabon, Boris Chidlovskii, and Jerome Revaud.
\newblock {DUSt3R: Geometric 3D Vision Made Easy}, 2023.
\newblock URL \url{https://arxiv.org/abs/2312.14132}.

\bibitem[Wang et~al.(2025{\natexlab{c}})Wang, Xu, Shan, Zhang, Diao, Duan,
  Huang, Liang, and Ma]{cinetechbench}
Xinran Wang, Songyu Xu, Xiangxuan Shan, Yuxuan Zhang, Muxi Diao, Xueyan Duan,
  Yanhua Huang, Kongming Liang, and Zhanyu Ma.
\newblock {CineTechBench: A Benchmark for Cinematographic Technique
  Understanding and Generation}, 2025{\natexlab{c}}.
\newblock URL \url{https://arxiv.org/abs/2505.15145}.

\bibitem[Wang et~al.(2025{\natexlab{d}})Wang, Li, Yan, He, Yu, Zeng, Wang, Ma,
  Huang, Gao, Dou, Chen, Wang, Qiao, Wang, and Wang]{internvideo25}
Yi~Wang, Xinhao Li, Ziang Yan, Yinan He, Jiashuo Yu, Xiangyu Zeng, Chenting
  Wang, Changlian Ma, Haian Huang, Jianfei Gao, Min Dou, Kai Chen, Wenhai Wang,
  Yu~Qiao, Yali Wang, and Limin Wang.
\newblock {InternVideo2.5: Empowering Video MLLMs with Long and Rich Context
  Modeling}, 2025{\natexlab{d}}.
\newblock URL \url{https://arxiv.org/abs/2501.12386}.

\bibitem[Wang et~al.(2025{\natexlab{e}})Wang, Zhou, Zhu, Chang, Zhou, Li, Chen,
  Pang, Shen, and He]{pi3}
Yifan Wang, Jianjun Zhou, Haoyi Zhu, Wenzheng Chang, Yang Zhou, Zizun Li, Junyi
  Chen, Jiangmiao Pang, Chunhua Shen, and Tong He.
\newblock {$\pi^3$: Permutation-Equivariant Visual Geometry Learning},
  2025{\natexlab{e}}.
\newblock URL \url{https://arxiv.org/abs/2507.13347}.

\bibitem[Wang et~al.(2026{\natexlab{b}})Wang, Ji, Cao, Shen, Xiao, Lyu, Xie,
  Liu, Tian, Long, Zhang, Cai, Chen, Zhao, Shi, Tang, Lyu, Tan, Zhang, Hu, Gao,
  Chen, Zhao, Xu, Zhu, Wang, Feng, Zhang, Zhao, Ao, Xie, Liu, Yao, Zhang, Liu,
  Zhang, Jiao, Yang, Wei, Liu, Pan, Nie, Men, Cui, Jin, Li, Luo, Mu, Wei, Yan,
  Zhao, Zheng, Li, Lin, Huang, Wang, and Wang]{orca}
Yihao Wang, Yuheng Ji, Mingyu Cao, Yanqing Shen, Runze Xiao, Huaihai Lyu,
  Senwei Xie, Euan Liu, Klara Tian, Tianfeng Long, Yichi Zhang, Zhengliang Cai,
  Ruike Chen, Jifan Zhao, Ruochuan Shi, Zihan Tang, Jing Lyu, Wenxing Tan,
  Ningbo Zhang, Yangtao Hu, Yuming Gao, Xiansheng Chen, Junkai Zhao, Congsheng
  Xu, Boan Zhu, Ziqi Wang, Yupu Feng, Qiongqiong Zhang, Yingli Zhao, Yulong Ao,
  Shaoxuan Xie, You Liu, Guocai Yao, Leiduo Zhang, Xiaodan Liu, Yunyan Zhang,
  Yance Jiao, Xinyan Yang, Jiaxing Wei, Xu~Liu, Tengfei Pan, Shaokai Nie,
  Chunlei Men, Sen Cui, Xiaojie Jin, Hongyang Li, Jianlan Luo, Yao Mu, Yunchao
  Wei, Jun Yan, Hang Zhao, Xiaolong Zheng, Jiaming Li, Yonghua Lin, Tiejun
  Huang, Zhongyuan Wang, and Pengwei Wang.
\newblock {Orca: The World is in Your Mind}, 2026{\natexlab{b}}.
\newblock URL \url{https://arxiv.org/abs/2606.30534}.

\bibitem[Wang et~al.(2026{\natexlab{c}})Wang, Yu, Rao, Ling, Li, Wang, Gao,
  Zhou, Liang, Liu, Zhang, Huang, Xu, Yuan, Yuan, Tan, Zhang, Huang, Zhang, Wu,
  Hu, and Zhang]{hyembodiedvlm}
Ziyi Wang, Xumin Yu, Yongming Rao, Yonggen Ling, Yunheng Li, Oran Wang, Mingqi
  Gao, Yuchen Zhou, Yves Liang, Zuyan Liu, Yani Zhang, Rui Huang, Xiaoran Xu,
  Bowen Yuan, Yifu Yuan, Xu~Tan, He~Zhang, Yufei Huang, Shenghao Zhang,
  Hongsheng Wu, Han Hu, and Zhengyou Zhang.
\newblock {Hy-Embodied-VLM-1.0: Efficient Physical-World Agents},
  2026{\natexlab{c}}.
\newblock URL \url{https://arxiv.org/abs/2607.12894}.

\bibitem[Wu et~al.(2025{\natexlab{a}})Wu, Liu, Hung, and Duan]{spatialmllm}
Diankun Wu, Fangfu Liu, Yi-Hsin Hung, and Yueqi Duan.
\newblock {Spatial-MLLM: Boosting MLLM Capabilities in Visual-based Spatial
  Intelligence}, 2025{\natexlab{a}}.
\newblock URL \url{https://arxiv.org/abs/2505.23747}.

\bibitem[Wu et~al.(2026)Wu, Cai, Li, Ge, Sun, Yuan, and Wang]{camreasoner}
Hang Wu, Yujun Cai, Zehao Li, Haonan Ge, Bowen Sun, Junsong Yuan, and Yiwei
  Wang.
\newblock {CamReasoner: Reinforcing Camera Movement Understanding via
  Structured Spatial Reasoning}, 2026.
\newblock URL \url{https://arxiv.org/abs/2602.00181}.

\bibitem[Wu et~al.(2025{\natexlab{b}})Wu, Zheng, Zhou, and Lu]{point3r}
Yuqi Wu, Wenzhao Zheng, Jie Zhou, and Jiwen Lu.
\newblock {Point3R: Streaming 3D Reconstruction with Explicit Spatial Pointer
  Memory}, 2025{\natexlab{b}}.
\newblock URL \url{https://arxiv.org/abs/2507.02863}.

\bibitem[Yang et~al.(2025{\natexlab{a}})Yang, Sax, Liang, Henaff, Tang, Cao,
  Chai, Meier, and Feiszli]{fast3r}
Jianing Yang, Alexander Sax, Kevin~J. Liang, Mikael Henaff, Hao Tang, Ang Cao,
  Joyce Chai, Franziska Meier, and Matt Feiszli.
\newblock {Fast3R: Towards 3D Reconstruction of 1000+ Images in One Forward
  Pass}, 2025{\natexlab{a}}.
\newblock URL \url{https://arxiv.org/abs/2501.13928}.

\bibitem[Yang et~al.(2026)Yang, Zhao, Pan, Yang, Zhang, Kang, Xu, Li, and
  Xie]{cambrianp}
Jihan Yang, Zifan Zhao, Xichen Pan, Shusheng Yang, Junyi Zhang, Bingyi Kang,
  Hu~Xu, Shang-Wen Li, and Saining Xie.
\newblock {Cambrian-P: Pose-Grounded Video Understanding}, 2026.
\newblock URL \url{https://arxiv.org/abs/2605.22819}.

\bibitem[Yang et~al.(2025{\natexlab{b}})Yang, Yang, Huang, Brown, Yang, Yu,
  Tong, Zheng, Xu, Wang, Lu, Fergus, LeCun, Fei-Fei, and Xie]{cambrians}
Shusheng Yang, Jihan Yang, Pinzhi Huang, Ellis Brown, Zihao Yang, Yue Yu,
  Shengbang Tong, Zihan Zheng, Yifan Xu, Muhan Wang, Daohan Lu, Rob Fergus,
  Yann LeCun, Li~Fei-Fei, and Saining Xie.
\newblock {Cambrian-S: Towards Spatial Supersensing in Video},
  2025{\natexlab{b}}.
\newblock URL \url{https://arxiv.org/abs/2511.04670}.

\bibitem[Yu et~al.(2026)Yu, Qu, Wang, Zhu, and Ke]{depthvlm}
Hanxun Yu, Xuan Qu, Yuxin Wang, Jianke Zhu, and Lei Ke.
\newblock {Unlocking Dense Metric Depth Estimation in VLMs}, 2026.
\newblock URL \url{https://arxiv.org/abs/2605.15876}.

\bibitem[Yu et~al.(2025)Yu, Cheng, Zhou, Yang, and
  Wang]{yu2025rgbonlygaussiansplattingslam}
Sicheng Yu, Chong Cheng, Yifan Zhou, Xiaojun Yang, and Hao Wang.
\newblock {RGB-Only Gaussian Splatting SLAM for Unbounded Outdoor Scenes},
  2025.
\newblock URL \url{https://arxiv.org/abs/2502.15633}.

\bibitem[Yue et~al.(2023)Yue, Ni, Zhang, Zheng, Liu, Zhang, Stevens, Jiang,
  Ren, Sun, Wei, Yu, Yuan, Sun, Yin, Zheng, Yang, Liu, Huang, Sun, Su, and
  Chen]{mmmu}
Xiang Yue, Yuansheng Ni, Kai Zhang, Tianyu Zheng, Ruoqi Liu, Ge~Zhang, Samuel
  Stevens, Dongfu Jiang, Weiming Ren, Yuxuan Sun, Cong Wei, Botao Yu, Ruibin
  Yuan, Renliang Sun, Ming Yin, Boyuan Zheng, Zhenzhu Yang, Yibo Liu, Wenhao
  Huang, Huan Sun, Yu~Su, and Wenhu Chen.
\newblock {MMMU: A Massive Multi-discipline Multimodal Understanding and
  Reasoning Benchmark for Expert AGI}, 2023.
\newblock URL \url{https://arxiv.org/abs/2311.16502}.

\bibitem[Zhang et~al.(2024)Zhang, Herrmann, Hur, Jampani, Darrell, Cole, Sun,
  and Yang]{monst3r}
Junyi Zhang, Charles Herrmann, Junhwa Hur, Varun Jampani, Trevor Darrell,
  Forrester Cole, Deqing Sun, and Ming-Hsuan Yang.
\newblock {MonST3R: A Simple Approach for Estimating Geometry in the Presence
  of Motion}, 2024.
\newblock URL \url{https://arxiv.org/abs/2410.03825}.

\bibitem[Zhang et~al.(2025)Zhang, Wang, Xu, Xue, Rupprecht, Zhou, Shen, and
  Wetzstein]{flare}
Shangzhan Zhang, Jianyuan Wang, Yinghao Xu, Nan Xue, Christian Rupprecht,
  Xiaowei Zhou, Yujun Shen, and Gordon Wetzstein.
\newblock {FLARE: Feed-forward Geometry, Appearance and Camera Estimation from
  Uncalibrated Sparse Views}, 2025.
\newblock URL \url{https://arxiv.org/abs/2502.12138}.

\bibitem[Zhao et~al.(2025)Zhao, Zhang, Xu, Chang, Chen, Li, Sun, and
  Wei]{spacemind}
Ruosen Zhao, Zhikang Zhang, Jialei Xu, Jiahao Chang, Dong Chen, Lingyun Li,
  Weijian Sun, and Zizhuang Wei.
\newblock {SpaceMind: Camera-Guided Modality Fusion for Spatial Reasoning in
  Vision-Language Models}, 2025.
\newblock URL \url{https://arxiv.org/abs/2511.23075}.

\bibitem[Zheng et~al.(2026)Zheng, Zhang, Zhang, Huang, Kumar, Kong, Tuzel, Li,
  and Krishna]{trajtok}
Chenhao Zheng, Jieyu Zhang, Jianing Zhang, Weikai Huang, Ashutosh Kumar, Quan
  Kong, Oncel Tuzel, Chun-Liang Li, and Ranjay Krishna.
\newblock {TrajTok: Learning Trajectory Tokens enables better Video
  Understanding}, 2026.
\newblock URL \url{https://arxiv.org/abs/2602.22779}.

\bibitem[Zhuo et~al.(2025)Zhuo, Zheng, Guo, Wu, Zhou, and Lu]{streamvggt}
Dong Zhuo, Wenzhao Zheng, Jiahe Guo, Yuqi Wu, Jie Zhou, and Jiwen Lu.
\newblock {Streaming 4D Visual Geometry Transformer}, 2025.
\newblock URL \url{https://arxiv.org/abs/2507.11539}.

\end{thebibliography}
